\documentclass[10pt,twocolumn,letterpaper]{article}

\usepackage{cvpr}              % To produce the CAMERA-READY version

\usepackage{times}
\usepackage{epsfig}
\usepackage{graphicx}
\usepackage{amsmath}
\usepackage{amssymb}
\usepackage[dvipsnames]{xcolor}
\usepackage{subcaption}
\usepackage{caption}
\usepackage[labelfont=bf]{caption}
\usepackage{float}
\usepackage{booktabs}
\usepackage{colortbl}
\usepackage{adjustbox}
\usepackage{microtype}
\usepackage{multirow}
\usepackage{pifont}
\usepackage{xcolor}
\newcommand{\xmark}{\ding{55}}%
\newcommand*\colourcheck[1]{%
  \expandafter\newcommand\csname #1check\endcsname{\textcolor{#1}{\ding{52}}}}
\colourcheck{blue}
\colourcheck{green}
\colourcheck{red}

\usepackage{array}

\usepackage[percent]{overpic}
\usepackage{pgfplots}
\pgfplotsset{compat=newest}
\definecolor{customTeal}{HTML}{1a9696}
\definecolor{customgrey}{HTML}{47494b}
 
\definecolor{custom_Turquoise}{HTML}{00897B}
\definecolor{custom_RoyalBlue}{HTML}{1E88E5}
\definecolor{custom_Rust}{HTML}{795548}
\definecolor{custom_SlateGray}{HTML}{546E7A}
\definecolor{custom_Gold}{HTML}{C48A00}
\definecolor{custom_NavyBlue}{HTML}{303F9F}
\definecolor{custom_Crimson}{HTML}{C20900}
\definecolor{custom_Purple}{HTML}{8E24AA}
\definecolor{custom_Lavender}{HTML}{E443FF}

\usepackage[dvipsnames]{xcolor}
\newcommand{\red}[1]{{\color{red}#1}}

\definecolor{cvprblue}{rgb}{0.21,0.49,0.74}
\usepackage[pagebackref,breaklinks,colorlinks,citecolor=cvprblue]{hyperref}

\def\paperID{10179} % *** Enter the Paper ID here
\def\confName{CVPR}
\def\confYear{2024}

\title{Sparse Semi-DETR: Sparse Learnable Queries for \\Semi-Supervised Object Detection}
\author{Tahira Shehzadi\(^{1,2,3}\), \hspace{-8mm}
\and
 Khurram Azeem Hashmi\(^{1,2,3}\),\hspace{-8mm}
\and
 Didier Stricker\(^{1,2,3}\),\hspace{-8mm}
\and
Muhammad Zeshan Afzal\(^{1,2,3}\) 
\and
{\(^{1}\)DFKI, \hspace{5mm}
\(^{2}\)RPTU Kaiserslautern-Landau, \hspace{5mm}
\(^{3}\)MindGarage-RPTU}\\ 
}
\begin{document}
\twocolumn[{
\renewcommand\twocolumn[1][]{#1}%
\maketitle
\begin{center}
    \centering
    \captionsetup{type=figure}
    \includegraphics[width=\textwidth]{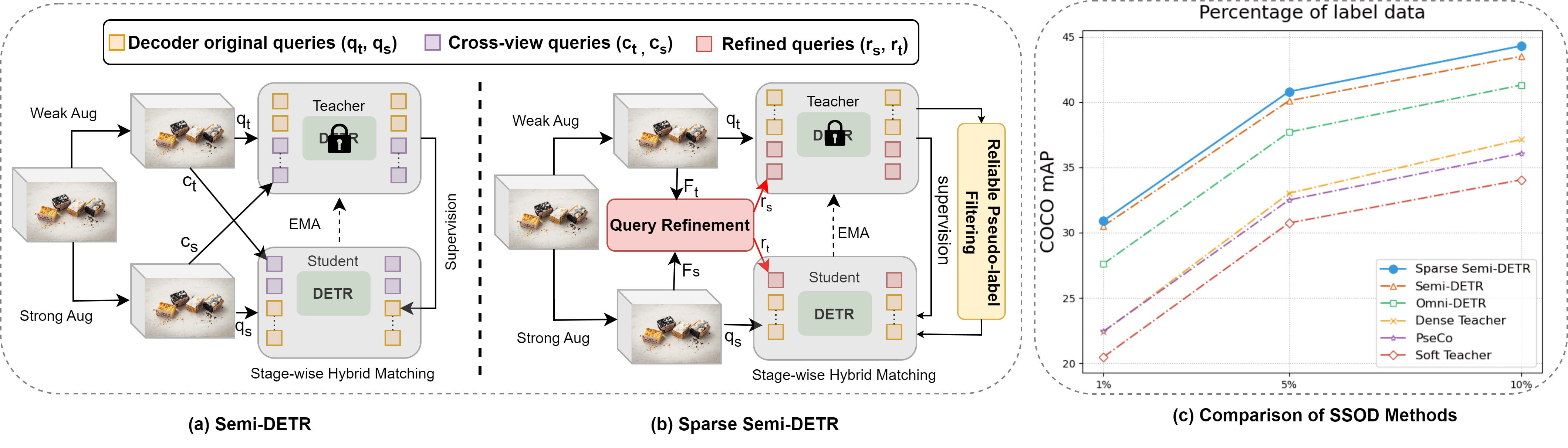}
    \captionof{figure}{\textbf{(a)-(b) Comparative Overview of SSOD advancements:} Sparse Semi-DETR's major improvement lies in its Query Refinement Module and Reliable Pseudo-label Filtering Module, significantly enhancing detection of small or obscured objects and reliability in complex scenarios, surpassing all other methods as shown in the graph in \textbf{(c)}. }
    \label{fig:intro_figure}
\end{center}%
}]
%\maketitlesupplementary
\maketitle
\begin{abstract}
In this paper, we address the limitations of the DETR-based semi-supervised object detection (SSOD) framework, particularly focusing on the challenges posed by the quality of object queries. In DETR-based SSOD, the one-to-one assignment strategy provides inaccurate pseudo-labels, while the one-to-many assignments strategy leads to overlapping predictions. These issues compromise training efficiency and degrade model performance, especially in detecting small or occluded objects. We introduce Sparse Semi-DETR, a novel transformer-based, end-to-end semi-supervised object detection solution to overcome these challenges. Sparse Semi-DETR incorporates a \textbf{Query Refinement Module} to enhance the quality of object queries, significantly improving detection capabilities for small and partially obscured objects.
Additionally, we integrate a \textbf{Reliable Pseudo-Label Filtering Module} that selectively filters high-quality pseudo-labels, thereby enhancing detection accuracy and consistency. On the MS-COCO and Pascal VOC object detection benchmarks, Sparse Semi-DETR achieves a significant improvement over current state-of-the-art methods that highlight Sparse Semi-DETR's effectiveness in semi-supervised object detection, particularly in challenging scenarios involving small or partially obscured objects.
%%We analyze the DETR-based framework and observe that quality and quantity of object queries significantly influence training efficiency and model performance. DETR-based semi-supervised object detection (SSOD) struggles with one-to-one assignments causing mismatches due to inaccurate pseudo labels and one-to-many assignments leading to overlaps, both affecting training efficiency and performance drop. Focusing on the quantity of object queries, the integration of one-to-many and one-to-one assignments in Semi-DETR complicates training and decrease performance, posing a challenge to the conventional optimization process in detection methods. (1) We introduce Sparse Semi-DETR, a transformer-based, end-to-end semi-supervised object detection solution designed to address these challenges. Specifically, we propose a Label Refinement module that improve quality of object queries and boost the model’s performance in detecting small objects and partially obscured objects. Additionally, we introduce the Reliable Pseudo-Label Filtering Module that filter out more high-quality pseudo-labels, enhancing detection accuracy and consistency. Sparse Semi-DETR achieves 44.3 mAP with ResNet-50 backbone using just 10\% labels on MS-COCO dataset, which surpasses the previous baselines by 0.8 mAP. When trained on fully annotated MS-COCO with additional unlabeled data, the performance further 2.1 increases 49.2 to 51.3 mAP, highlighting the effectiveness of our Sparse Semi-DETR. 
%Semi-supervised object detection aims to  impressive results using pseudo-labels when data is scarce. 
\end{abstract}    
\section{Introduction}
\label{sec:intorduction}
%Semi-Supervised Object Detection (SSOD) aims to enhance the performance of fully-supervised object detection by utilizing a large amount of unlabeled data
Semi-Supervised Object Detection (SSOD) aims to improve the effectiveness of fully supervised object detection through the integration of abundant unlabeled data~\cite{STAC,Self_Training_SSOD_CVPR20,SoftTeacher,Mean_Teachers_CVPR21,Instant-Teaching_CVPR21,Consistent_Teacher_CVPR23,DSL_CVPR22,SED_CVPR22,CSL_NIPS19,PseCo_ECCV22,UnbiasedTeacherv2_CVPR22}. It has applications in diverse fields, ranging from autonomous vehicles~\cite{autonomous_driving_icra22,semi_autonomous_driving_cvpr22} to healthcare~\cite{semi_medical_CVPR20,semi_medical_ECTC22}, where obtaining extensive labeled datasets is often impractical or cost-prohibitive~\cite{label_cost_PMLR23}.

Several SSOD methods~\cite{STAC,Self_Training_SSOD_CVPR20,SoftTeacher,Mean_Teachers_CVPR21,Instant-Teaching_CVPR21,Consistent_Teacher_CVPR23,DSL_CVPR22,SED_CVPR22,CSL_NIPS19,PseCo_ECCV22,UnbiasedTeacherv2_CVPR22} have been proposed. Two prevalent approaches in this domain are pseudo-labeling~\cite{STAC,unbiased_Teacher_ICLR21,Adapting_SVCI_AAAI23,Double_Check_IJCAI22,Self_Training_SSOD_CVPR20,SoftTeacher,Mean_Teachers_CVPR21,Instant-Teaching_CVPR21} and consistency-based regularization~\cite{Consistent_Teacher_CVPR23,DSL_CVPR22,SED_CVPR22,CSL_NIPS19,PseCo_ECCV22,UnbiasedTeacherv2_CVPR22}. STAC~\cite{STAC} introduced a simple multi-stage SSOD training method with pseudo-labeling and consistency training, later simplified by a Teacher-Student framework for generating pseudo-labels~\cite{ unbiased_Teacher_ICLR21}. Based on this framework, considerable research efforts have been directed towards enhancing the quality of pseudo-labels~\cite{SoftTeacher,Instant-Teaching_CVPR21}. These traditional SSOD methods are built upon conventional detectors like one-stage~\cite{yolov3,FCOS} and two-stage~\cite{fast_rcnn_Girshick_2015_ICCV,FasterRCNN}, which involve various manually designed components such as anchor boxes and non-maximum suppression
(NMS). Employing object detection methods in SSOD poses several potential challenges that must be carefully dealt with to obtain reasonable performance. These factors include overfitting of the labeled data~\cite{overfitting_ACM}, pseudo-label noise~\cite{noisy_labels_axiv}, bias induced through label imbalance~\cite{bias_nois_acm,revisiting_class_imbalance34}, and poor detection performance on small objects~\cite{Semi-DETR_cvpr23}. Recently, DETR-based~\cite{DETR,deformable_detr,dab89,dn42,dino_detr_ICLR_23,HDETR_CVPR23,OD_Transformers} SSOD methods~\cite{omni-DETR_cvpr22, Semi-DETR_cvpr23} remove the need for traditional components like NMS.

Even though DETR-based SSOD~\cite{omni-DETR_cvpr22,Semi-DETR_cvpr23} has progressed remarkably, state-of-the-art methods possess some limitations. (1) DETR-based SSOD methods perform poorly in the detection of small objects, as shown in Figure~\ref{fig:result1}. This is because these methods don't use multi-scale features~\cite{sparsedetr} like Feature Pyramid Networks (FPN)~\cite{fpn6}, which play an important role in identifying smaller objects as in CNN-based SSOD methods~\cite{SoftTeacher,Mean_Teachers_CVPR21,Instant-Teaching_CVPR21,Consistent_Teacher_CVPR23,DSL_CVPR22,SED_CVPR22,CSL_NIPS19,PseCo_ECCV22,UnbiasedTeacherv2_CVPR22}. Although recent advancements in DETR-based object detection~\cite{DETR,deformable_detr,dab89,dn42,dino_detr_ICLR_23,HDETR_CVPR23,OD_Transformers} have improved the detection of small objects, their SSOD adaptation is still unable to cater this challenge effectively~\cite{Semi-DETR_cvpr23}. (2) SSOD approaches~\cite{SoftTeacher,Mean_Teachers_CVPR21,Instant-Teaching_CVPR21,Semi-DETR_cvpr23} rely on handcrafted post-processing methods such as NMS~\cite{FasterRCNN}. This problem specifically appears in DETR-based SSOD when we use a large number of object queries and the one-to-many assignment strategy~\cite{Semi-DETR_cvpr23}. In DETR-based SSOD methods, this problem is partially solved using the one-to-one or hybrid (combination of one-to-one and one-to-many) assignment strategy. However, the hybrid assignment strategy is preferred because the one-to-one assignment strategy produces inaccurate pseudo-labels~\cite{omni-DETR_cvpr22}, thus resulting in inefficient learning. Although the number of duplicate bounding boxes is less in the hybrid strategy~\cite{Semi-DETR_cvpr23}, the amount is high enough to impact object detection performance adversely, as depicted in Figure~\ref{fig:result2}. (3) The pseudo-label generation produces both high and low-quality labels. The DETR-based SSOD methods lack an effective refinement strategy for one-to-many assignments, which is crucial for filtering out low-quality proposals.
%The DETR-based SSOD methods lack the refinement strategy in one-to-many assignment startegy  for filtering out low-quality proposals. 

To address the above mentioned issues, we propose enhancing the state-of-the-art DETR-based SSOD approach, namely 'Sparse Semi-DETR', presented
in Figure~\ref{fig:intro_figure} (b). Our approach involves expanding its architecture by integrating a couple of novel modules designed to mitigate the identified shortcomings. The key module among these is the \textbf{Query-Refinement module}, as depicted in Figure~\ref{fig:semi} and explained in Figure~\ref{fig:queries}.
This module significantly improves the quality of the queries and reduces their numbers. The proposed module uses the low-level features from the backbone and high-level features extracted directly from weakly augmented images using ROI alignment~\cite{Mask_RCNN_ICCV_17}. Fusing these features results in overcoming the first shortcoming, i.e., detecting small and obscured objects, as shown in Figure~\ref{fig:result1}. The attention mechanism drives the aggregation of the features, resulting in refined, high-quality features to carry forward. To ensure the quality of the query features, the attention mechanism is accompanied by a query-matching strategy for filtering irrelevant queries. Thus, the Query Refinement Module not only improves the quality of the queries but also reduces their numbers, giving rise to efficient processing. This module results in significantly fewer overlapping proposals, improving the performance overall, thereby solving the second limitation.
Besides, we introduce a \textbf{Reliable Pseudo-Label Filtering Module}, as illustrated in Figure~\ref{fig:semi}, inspired by Hybrid-DETR~\cite{HDETR_CVPR23} to address the third limitation. Employing this module significantly reduces the low-quality pseudo-labels. Therefore, it further reduces the amount of duplicate predictions that may still occur after the second stage of the hybrid assignment strategy. Our approach provides better results than previous SSOD methods, as shown in Figure~\ref{fig:intro_figure} (c).
The key contributions of this work can be outlined as follows:
\begin{enumerate}
    \item We present Sparse Semi-DETR, a novel approach in semi-supervised object detection, introducing two novel contributions. To our knowledge, we are the first to examine and propose query refinement and low-quality proposal filtering for the one-to-many query assignment strategy.  
    
    \item We introduce a novel query refinement module designed to improve object query features, particularly in complex detection scenarios such as identifying small or partially obscured objects. This enhancement not only boosts performance but also aids in learning semantic feature invariance among object queries.
    \item We introduce a Reliable Pseudo-Label Filtering Module specifically designed to reduce the effect of noisy pseudo-labels. This module is designed to efficiently identify and extract reliable pseudo boxes from unlabeled data using augmented ground truths, enhancing the consistency of the learning process.
    %\item On the MS-COCO and Pascal VOC object detection benchmarks, Sparse Semi-DETR achieves significant improvements in performance over current state-of-the-art methods. Our evaluations on the MS-COCO dataset demonstrate that Sparse Semi-DETR, with only 10\% labeled data and a ResNet-50 backbone, achieves a mean Average Precision (mAP) of 44.3, surpassing previous baselines by 0.8 mAP. Our method also demonstrates strong performance with a relatively large amount of labeled data. Specifically, When trained on the full COCO training set with additional unlabeled data, the model further improves, achieving a notable increase from 49.2 to 51.3 mAP.
    \item Sparse Semi-DETR outperforms current state-of-the-art methods on MS-COCO and Pascal VOC benchmarks. With only 10\% labeled data from MS-COCO using ResNet-50 backbone, it achieves a 44.3 mAP, exceeding prior baselines by 0.8 mAP. Additionally, when trained on the complete COCO set with extra unlabeled data, it further improves, rising from 49.2 to 51.3 mAP.
\end{enumerate}
\section{Related Work}
\label{sec:related_work}
\subsection{Object Detection} 
Object detection identifies and locates objects in images or videos. Deep learning-based object detection approaches are typically categorized into two primary groups: two-stage detectors~\cite{FasterRCNN,fast_rcnn_Girshick_2015_ICCV} and one-stage detectors~\cite{yolo45,SSD,retinaNet,FCOS}. These methods depend on numerous heuristics, such as generating anchors and NMS. Recently, DEtection TRansformer (DETR)~\cite{DETR} considers object detection as a set prediction problem, using transformer~\cite{attn_all_need_Neurips_17} to adeptly transform sparse object candidates~\cite{SparseRCNN} into precise target objects. Our Sparse Semi-DETR detects small or partially obscured objects in the DETR-based SSOD setting. Notably, our framework is compatible with various DETR-based detectors~\cite{deformable_detr,UPDETR_CVPR20,smca23,CondDE,WBdetr4,pnp6,yolos6,fpdetr,rego2}, offering flexibility in integration. 

\subsection{Semi-Supervised Object Detection} 
Most research in SSOD employs detectors categorized into three types: one-stage, two-stage, and DETR-based systems.

\noindent\textbf{One-stage}
STAC~\cite{STAC}, an early SSOD, introduced a simple training strategy combining pseudo-labeling and consistency training, later streamlined by a student-teacher framework for easier pseudo-label generation~\cite{ unbiased_Teacher_ICLR21}. DSL~\cite{DSL_CVPR22} introduced novel techniques including Adaptive Filtering, Aggregated Teacher, and uncertainty-consistency-regularization for improved generalization. Dense Teacher~\cite{DenseTeacher} introduced Dense Pseudo-Labels (DPL) for richer information and a region selection method to reduce noise.
%DSL~\cite{DSL_CVPR22} introduced the first anchor-free SSOD approach based on dense learning, featuring an adaptive filtering method and uncertainty regularization, achieving state-of-the-art results. Dense Teacher~\cite{DenseTeacher} is designed to utilize dense predictions from its teacher network as pseudo-labels, bypassing the threshold selection process.

\noindent\textbf{Two-stage.}~Humble Teacher~\cite{HumbleTeachers_CVPR21} uses soft labels and a teacher ensemble to boost pseudo-label reliability, matching other results. Instant-Teaching~\cite{Instant-Teaching_CVPR21} creates pseudo annotations from weak augmentations, treating them as ground truth under strong augmentations with Mixup~\cite{mixup_17}. Unbiased Teacher~\cite{unbiased_Teacher_ICLR21} tackles class imbalance in pseudo-labeling with focal loss, focusing on underrepresented classes. Soft Teacher~\cite{SoftTeacher} minimizes incorrect foreground proposal classification by applying teacher-provided confidence scores to reduce classification loss. PseCo~\cite{PseCo_ECCV22} enhances detector performance by combining pseudo-labeling with label and feature consistency methods, also using focal loss to address class imbalance.

\noindent\textbf{DETR-based.}~Omni-DETR~\cite{omni-DETR_cvpr22} is designed for omni-supervised detection and adapts to SSOD with a basic pseudo-label filtering method. It employs the one-to-one assignment strategy proposed in DETR~\cite{DETR}, and encounters challenges when dealing with inaccurate pseudo-bounding boxes produced by the teacher network. These inaccuracies result in reduced performance, highlighting its limitations. Semi-DETR~\cite{Semi-DETR_cvpr23} adopts a stage-wise strategy, employing a one-to-many matching strategy in the first stage and switching to a one-to-one matching strategy in the second stage. This approach provides NMS-free end-to-end detection benefits but reduces performance compared to a one-to-many assignment strategy. Moreover, omni-DETR and Semi-DETR struggle to detect small or occluded objects. Our work introduces an advanced query refinement module that significantly refines object queries, enhancing training efficiency and performance and leading to the detection of small or densely packed objects in the DETR-based SSOD framework.

\section{Preliminary}
In DETR-based SSOD, one-to-one assignment strategy, denoted by $\hat{\sigma}_{one2one}$, is achieved by applying Hungarian algorithm between the predictions made by the student model and the pseudo-labels provided by the teacher model as follows:
\begin{equation}
\hat{\sigma}_{one2one} = \underset{\sigma \in \mathcal{\xi }_N}{\arg\min} \sum_{j}^{N} \mathcal{L}_{\text{match}} \left( \hat{y}_j^{t}, \hat{y}_{\sigma(j)}^{s} \right)
\end{equation}
where $\mathcal{L}_{\text{match}}\left(\hat{y}_j^{t}, \hat{y}_{\sigma(j)}^{s}\right)$ is the matching cost between the pseudo-labels $\hat{y}_j^{t}$ generated by the teacher network and the predictions of the student network with index $\sigma(j)$ and $\mathcal{\xi}_N$ is the permutation of $N$ elements. Semi-DETR~\cite{Semi-DETR_cvpr23} addresses the issue of imprecise initial pseudo-labels by shifting from a one-to-one to a one-to-many assignment strategy, increasing the number of positive object queries to improve detection accuracy:
\begin{equation}
\hat{\sigma}_{one2many} = \left\{ \underset{\sigma_j \in C_M^N}{\arg\min} \sum_{k}^{M} \mathcal{L}_{\text{match}} \left( \hat{y}_j^{t}, \hat{y}_{\sigma_j(k)}^{s} \right) \right\}^{|\hat{y}^t|}_j
\end{equation}
%where \( C^M_N \) is the combination of \( M \) and \( N \), which indicates that a subset of \( M \) proposals is matched with each pseudo box \( \hat{y}^t_j \). 
where \( C^M_N \) represents the combination of \( M \) and \( N \), denoting that a subset of \( M \) proposals is associated with each pseudo box \( \hat{y}^t_j \). Semi-DETR initially adopts a one-to-many assignment to improve label quality, then shifts to one-to-one assignment for an NMS-free model. This approach adopts a one-to-many assignment strategy aimed at boosting performance, but it's less effective with small or occluded objects.

%-------------------------------------------------------------------------

\section{Sparse Semi-DETR}
\label{sec:method}
\begin{figure*}
\centering
\includegraphics[width=.90\linewidth]{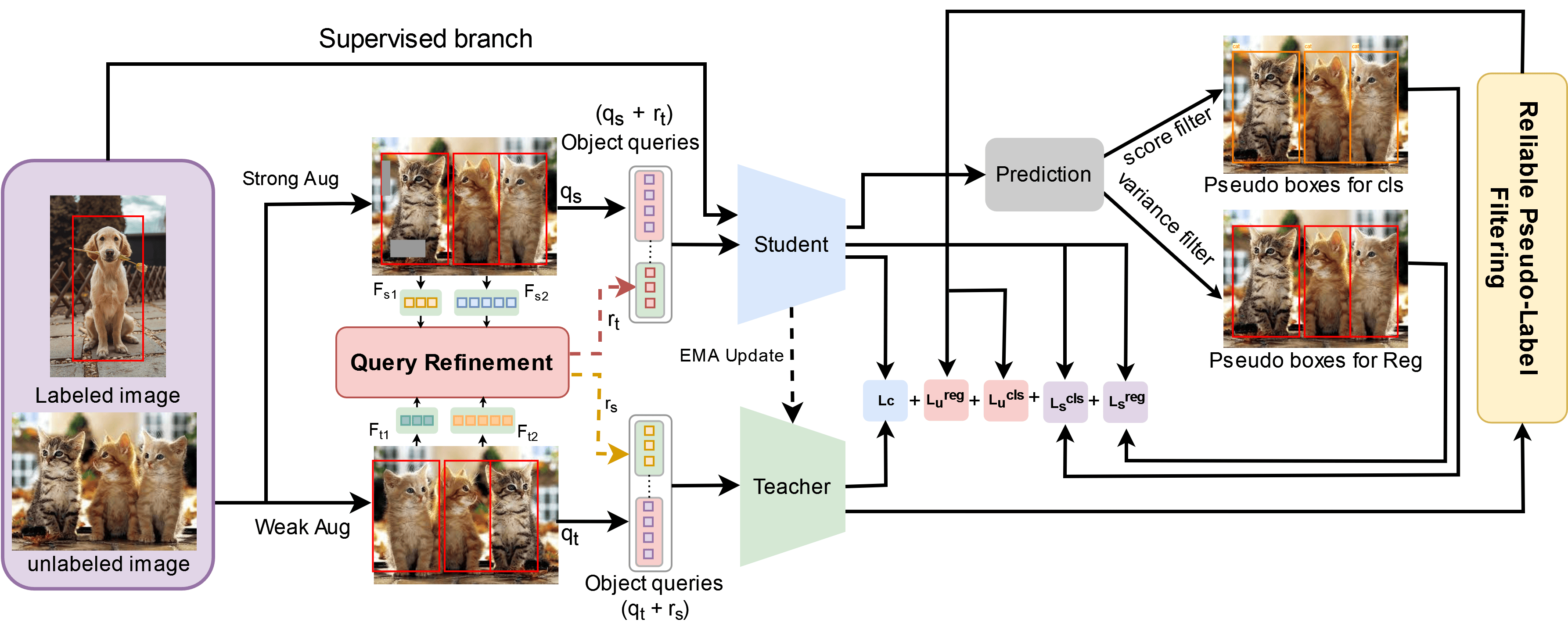}
   \caption{\textbf{An overview of the Sparse Semi-DETR framework.} It contains two networks: the student network and the teacher network. Labeled data is used for student network training, employing a supervised loss. Unlabeled data is fed to the teacher network with weak augmentation and the student network with strong augmentation. The teacher network takes unlabeled data to generate pseudo-labels. Here, the query refinement module provides refined queries to avoid incorrect bipartite matching with teacher-generated pseudo-labels. For a detailed overview of the query refinement module, see Figure~\ref{fig:queries}.  Furthermore, a  Reliable Pseudo-label Filtering strategy is employed to filter low-quality pseudo-labels progressively during training.}
\label{fig:semi}
\vspace{-10pt}
\end{figure*} 
In semi-supervised learning, a collection of labeled data denoted as $D_l$, where $D_l = {\{ x_{i}^l, y_{i}^l \}}_{i=1}^{N_l}$ is given, along with a set of unlabeled data represented as $D_u$, where 
$D_u = \{{x_{i}^u}\}_{i=1}^{N_u}$. Here, $N_l$ and $N_u$ correspond to the number of labeled and unlabeled data. The annotations $y_{i}^l$ for the label data $x^l$ contain object labels and bounding box information. The pipeline of the Sparse Semi-DETR framework is depicted in Figure~\ref{fig:semi}. It introduces a Query Refinement Module for processing query features to enhance semantic representation for complex detection scenarios, such as identifying small or partially obscured objects. Additionally, we integrate a Reliable Pseudo-Label Filtering Module that selectively filters high-quality pseudo-labels, thereby enhancing detection accuracy. For comparison purposes, we employ DINO~\cite{dino_detr_ICLR_23} with a ResNet-50 backbone. This section gives a detailed overview of the modules of Sparse Semi-DETR. We explain briefly our semi-supervised approach in Appendix~\textcolor{red}{A1.1}.

\subsection{Query Refinement} 
\begin{figure}
\centering
\includegraphics[width=\linewidth]{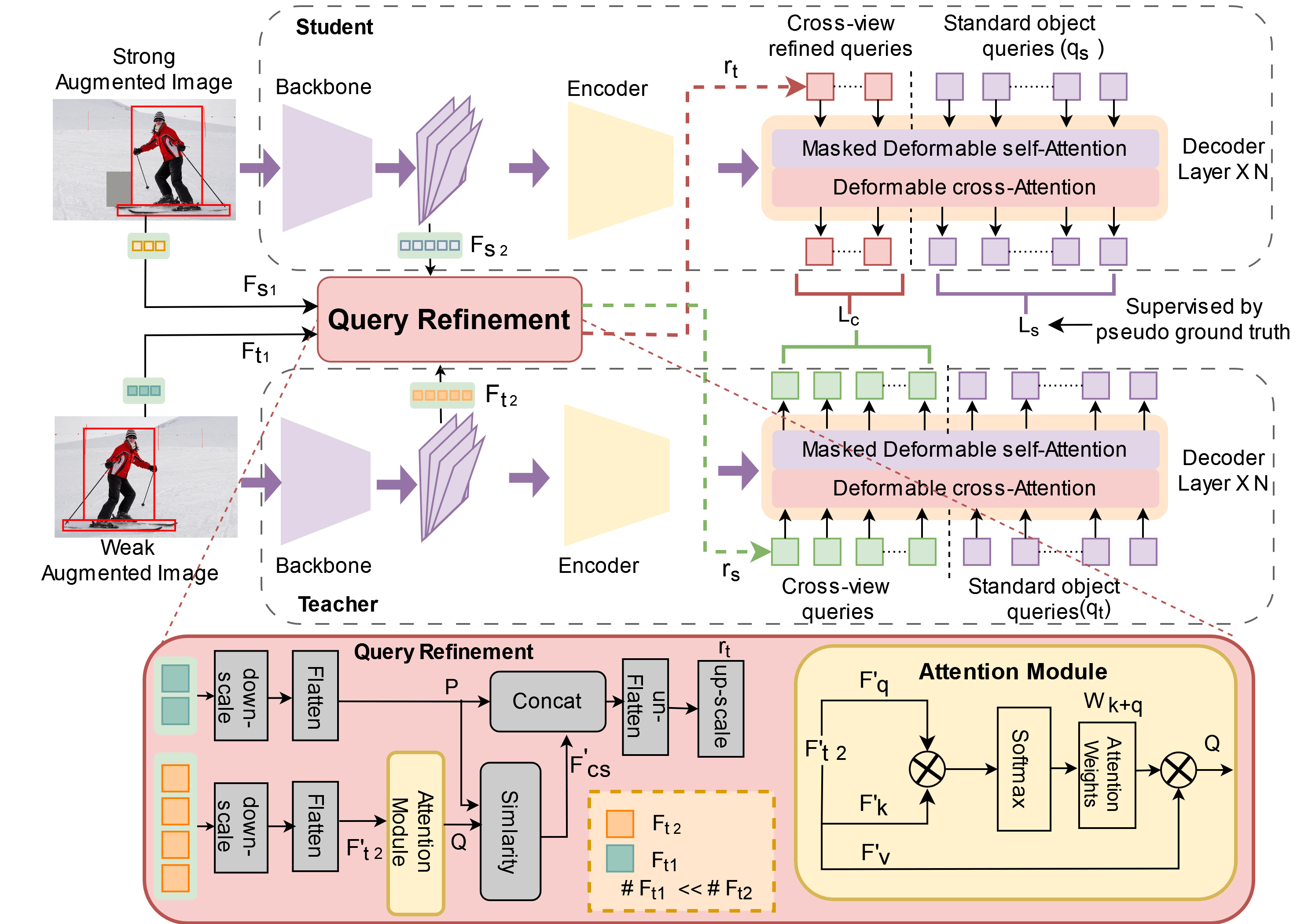}
   \caption{ \textbf{Overview of the Query Refinement Module.} The query features from strong and weak augmented unlabeled images are refined through the Query Refinement Module. It amplifies the semantic representation of object queries and improves performance for small objects. For the best view, zoom in.}
\label{fig:queries}
\vspace{-10pt}
\end{figure}

Inspired by recent advancements in vision-based networks~\cite{Zero_reference_CVPR20,FE_ECCV22}, we introduce an innovative approach to enhance object query features. For each unlabeled image \( I \in \mathbb{R}^{H \times W \times C} \), we extract query features \( F_{s1} \in \mathbb{R}^{b \times W_1 \times 256} \) from strongly augmented image \( I \). 
Similarly, we extract query features \( F_{t1} \) from weakly augmented image \( I \), also in the same dimension. Subsequently, feature extraction from the image backbones occurs. This results in the generation of features \( F_{s2} \) for the student and \( F_{t2} \) for the teacher network as \(  \mathbb{R}^{b \times W_2 \times 256} \). These features, encompassing both label and bounding box details, vary with the batch size, as indicated by \( b \). The feature sets \( W_1 \) and \( W_2 \) differ in size, with \( W_2 \) being substantially larger than \( W_1 \). We provide a brief overview of each component of Query Refinement Module as illustrated in Figure~\ref{fig:queries}.

\newcommand{\thickbar}[1]{\mathbf{\bar{\vphantom{#1}#1}}}
\noindent\textbf{Query Refinement Module.}
In our approach, we handle multi-scale features \(F_{t1}\) and \(F_{t2}\) with a focus on effective aggregation. The finer details are encapsulated within the features \(F_{t1}\), while the features \(F_{t2}\) encapsulate more abstract elements such as shapes and patterns. Simple aggregation of these features has been shown to degrade performance, as indicated in Table~\ref{tab:attention_queries_h}). To solve this issue, we implement dual strategies to extract local and  global information from high and low-resolution features. High-resolution features are crucial for detecting small objects. However, processing them with attentional operations is computationally demanding. To address this, we firstly convert the query label features \( F_{t2} \in (\mathbb{R}^{b \times W_2 \times 256 })\) into \( F'_{t2} \in (\mathbb{R}^{b \times W_2 \times 16 })\) by decreasing the channel dimension, and retaining the original resolution \(b\times W_2\).  Then, we apply attentional mechanism on \(F'_{t2}\) to calculate the attentional weights \( W_{k+q} \) in attention block as follows:
\begin{equation}
    \begin{aligned}
        W_{k+q} = F'_{k} \cdot F'_{q},
    \end{aligned}
    \label{eq:kq_product}
\end{equation}
\begin{equation}
    \begin{aligned}
       \thickbar{W}_{k+q} = \frac{\exp(W_{k+q})}{\sum_{l=1}^{L}\exp(W_{k+q})},
    \end{aligned}
    \label{eq:kq_softmax}
\end{equation}
where \( W_{k+q} \) is the attentional weights of \( F'_k \) and \( F'_q \), and $\thickbar{W}_{k+q}$ is the normalized form of \( W_{k+q} \). Using normalized attention weights, we compute the enhanced queries representation $Q$ as follows:
\begin{equation}
Q = \thickbar{W}_{k+q} \cdot F'_{v}
\end{equation}
now we find the similarity between the attentional \(F_{t2}\) features and \(F_{t1}\) features to obtain $F'_{cs}\in \mathbb{R}^{{b}\times{W_1} \times 16}$ from $ Q \in \mathbb{R}^{{b}\times{W_2} \times 16}$ as follows:
\begin{equation}
F'_{cs}= \frac{\sum_{i=l}^{n} P_l Q_l}{\sqrt{\sum_{l=1}^{n} P_l^2} \sqrt{\sum_{l=1}^{n} Q_l^2}}
\end{equation}
where $P$ and $Q$ are \(F_{t1}\) and attentional \(F_{t2}\) features, respectively. Then, we concatenate $F'_{cs}$ with $P$ to obtain refined query features. Interestingly, we observe a performance drop when our feature refinement strategy is applied to strongly augmented image features for the teacher network, as detailed in Table~\ref{tab:ST_attention_b}. However, we achieve optimal results by concatenating strongly augmented image features and applying our refinement strategy to weakly augmented image features. Consequently, we proceed by concatenating the features \( F_{s1} \) with \( F_{s2} \), thereby obtaining the query features \( r_{s} \). Note that \(r_{t}\), despite having a dimensional size equivalent to $F'_{cs}+ P$, encapsulates substantially more intricate representations. This improved performance is due to the integration of high-resolution and low-resolution features.

Then, we form the decoder queries in the student-teacher network by merging the teacher's original queries \( q_t \) with refined queries \( r_s \), and the student's original queries \( q_s \) with refined queries \( r_t \), respectively. This integration forms the inputs for the decoder as follows:
\begin{gather}
\hat{o}_t, {o}_t = \text{Dec}_t ([r_s, q_t], E_t[A]) \\ 
\hat{o}_s, {o}_s = \text{Dec}_s ([r_t, q_s], E_s[A])
\end{gather}
Where \( E_s \) and \( E_t \) refer to the encoded image features. \( \hat{o}_s \) and \( \hat{o}_t \) indicate the decoded features of refined queries, while \( o_s \) and \( o_t \) represent the decoded features of original object queries. Here, t is for teacher and s for student network. Following DN-DETR~\cite{dn42}, we use the attention mask \( A \) to protect information leakage, ensuring the learning process's integrity.
\subsection{Reliable Pseudo-Label Filtering Module} 
The one-to-many training strategy, while effective, causes duplication prediction in the first stage. We introduce a pseudo-label filtering module to address this and improve the filtering of pseudo boxes rich in semantic content for refined query learning. This module is designed to efficiently identify and extract reliable pseudo boxes from unlabeled data using augmented ground truths. We employ m groups of ground truths \(\hat{g} = \{\hat{g}_1, \hat{g}_2, \ldots, \hat{g}_{m}\}\) for one-to-many assignment strategy and select the top-k predictions as follows:
\begin{equation}
\hat{\sigma}_{one2many} = \left\{ \underset{\sigma_j \in C_M^N}{\arg\min} \sum_{k}^{M} \mathcal{L}_{\text{match}} \left( \hat{y}_j^{t}, \hat{g}_{\sigma_j(k)}^{s} \right) \right\}^{|\hat{y}^t|}_j
\end{equation}
%where \( C^M_N \) is the combination of \( M \) and \( N \), which indicates that a subset of \( M \) proposals is matched with each pseudo box \( \hat{y}^t_j \). 
where \( C^M_N \) represents the combination of \( M \) and \( N \), denoting that a subset of \( M \) proposals is associated with each pseudo box \( \hat{y}^t_j \). Here, m is set to 6. Furthermore, we use the remaining predictions to filter out duplicates in the top-k predictions in the one-to-one matching branch. Through this improved selection scheme, we achieve a performance improvement of 0.4 mAP when m is set to 6, as shown in Table~\ref{tab:effect_each_module_a}. However, we observe no significant benefits when increasing m greater than 6, as detailed in Table~\ref{tab:LR_m}.

\begin{figure*}
\centering
\includegraphics[width=\linewidth]{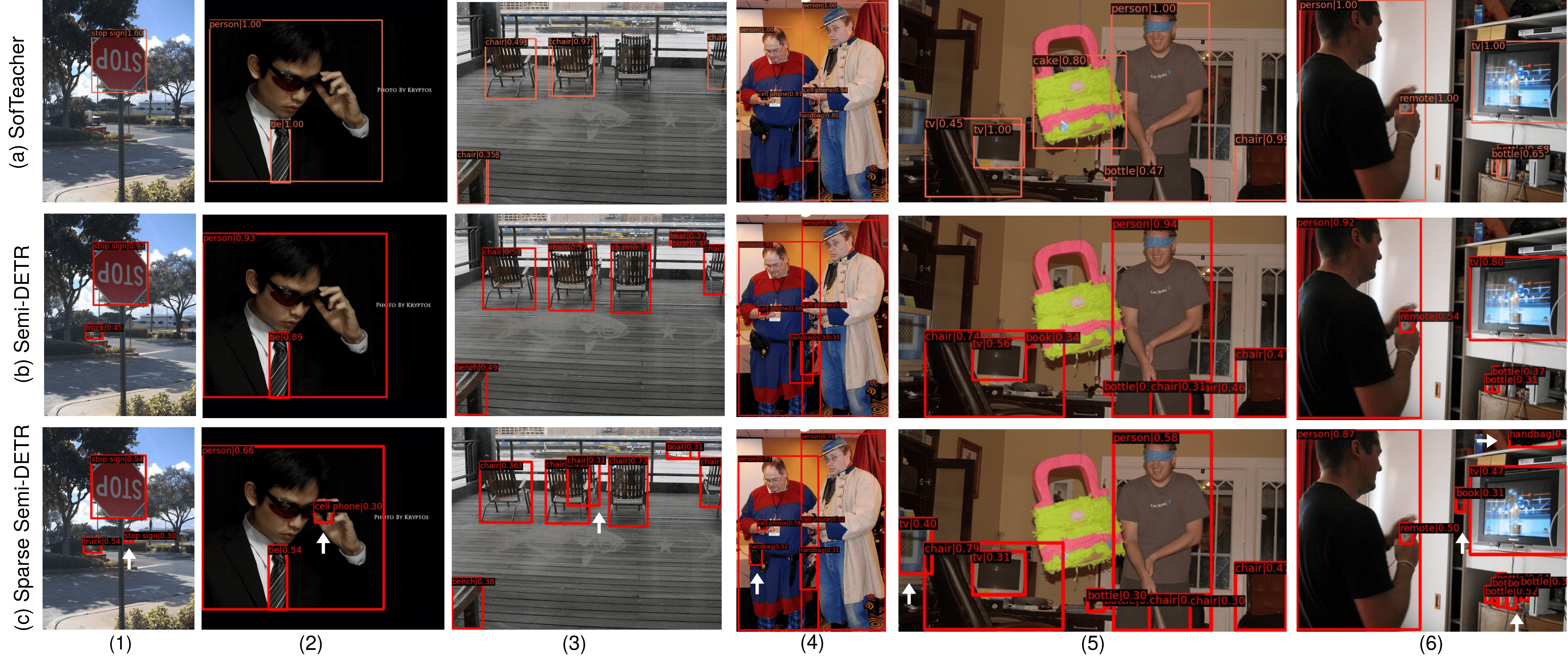}
   \caption{\textbf{Visual comparison of Sparse Semi-DETR with the
two previous approaches on the COCO 10\% label dataset.} These results highlight Sparse Semi-DETR's capabilities, particularly in identifying small objects and those obscured by obstacles (as indicated by white arrows) in the third-row images. For optimal clarity and detail, please zoom in.}
\label{fig:result1}
\vspace{-10pt}
\end{figure*} 
\section{Experiments}
\subsection{Datasets}
We evaluate our approach on the MS-COCO~\cite{coco14} and Pascal VOC~\cite{pascalvoc_15} datasets, benchmarking it against current SOTA SSOD methods. Following~\cite{SoftTeacher,Semi-DETR_cvpr23}, Sparse Semi-DETR is evaluated in three scenarios:~\textbf{COCO-Partial.} We use 1\%, 5\%, 10\% of \texttt{train2017} as label data and rest as unlabeled data.~\textbf{COCO-Full.} We take \texttt{train2017} as label data and \texttt{unlabel2017} as unlabel data.~\textbf{VOC.} We take \texttt{VOC2007} as label data and \texttt{VOC2012} as unlabel data. Evaluation metrics include $AP_{50:95}$, $AP_{50}$, and $AP_{75}$~\cite{SoftTeacher,Semi-DETR_cvpr23}.
\subsection{Implementation Details}
We set the quantity of DINO original object queries to 900. For the setting hyperparameters, following~\cite{Semi-DETR_cvpr23} : (1) In the COCO-Partial setting, we set the training iterations to 120k with a labeled to unlabeled data ratio of 1:4. The first 60k iterations adopt a one-to-many assignment strategy. (2) In the COCO-Full setting, Sparse Semi-DETR is trained for 240k iterations with labeled to unlabeled data ratio of 1:1. The first 120k iterations adopt a one-to-many assignment strategy. (3) In the VOC setting, we train the network for 60k iterations with a labeled to unlabeled data ratio of 1:4. The first 40k iterations adopt a one-to-many assignment strategy. For all experiments, the filtering threshold $\sigma$ value is 0.4. We set the value of m to 6 and the value of k to 4. We provide complete implementation details for each experiment in Appendix~\textcolor{red}{A1.2}.

\definecolor{lb}{RGB}{100,149,237}
\section{Results and Comparisons}
We evaluate Sparse Semi-DETR and compare it against current SOTA SSOD methods. Our results demonstrate the superior performance of Sparse Semi-DETR in these aspects: (1) its effectiveness compared to both one-stage and two-stage detectors, (2) its comparison with traditional DETR-based detectors, and (3) its exceptional proficiency in accurately detecting small and partially occluded objects. We provide more results details in Appendix~\textcolor{red}{A1.3}.
\begin{table*}
\begin{minipage}[t]{\linewidth}
\centering
\footnotesize
\begin{adjustbox}{width=.88\linewidth}
 \begin{tabular}{l c c c c}
  \toprule
    \multirow{2}{*}{Methods} &
    \multirow{2}{*}{Reference} &
    \multicolumn{3}{c}{COCO-Partial} \\
      % \vspace{2pt}
      \cmidrule(lr){3-5}   & &{ 1\% } & {5\%} &  { 10\%} \\
   \toprule 
   FCOS~\cite{FCOS} (Supervised) & - &  8.43 $\pm$ 0.03 & 17.01 $\pm$ 0.01 & 20.98 $\pm$ 0.01 \\
   DSL~\cite{DSL_CVPR22} & CVPR22  &  22.03 $\pm$ 0.28 \textcolor{lb}{(+13.98)} & 30.87 $\pm$ 0.24 \textcolor{lb}{(+13.86)} & 36.22 $\pm$ 0.18 \textcolor{lb}{(+15.24)} \\
    Unbiased Teacher v2~\cite{UnbiasedTeacherv2_CVPR22} & CVPR22&  22.71 $\pm$ 0.42 \textcolor{lb}{(+14.28)} & 30.08 $\pm$ 0.04 \textcolor{lb}{(+13.07)} & 32.61 $\pm$ 0.03 \textcolor{lb}{(+11.63)} \\
    Dense Teacher~\cite{DenseTeacher} & ECCV22 &  22.38 $\pm$ 0.31 \textcolor{lb}{(+13.95)} & 33.01 $\pm$ 0.14 \textcolor{lb}{(+16.00)} & 37.13 $\pm$ 0.12 \textcolor{lb}{(+16.15)} \\
    \midrule 
    Faster RCNN~\cite{FasterRCNN} (Supervised)  & - &  9.05 $\pm$ 0.16 & 18.47 $\pm$ 0.22 & 23.86 $\pm$ 0.81 \\
    Humble Teacher~\cite{HumbleTeachers_CVPR21} & CVPR22 & 16.96 $\pm$ 0.38 \textcolor{lb}{(+7.91)} & 27.70 $\pm$ 0.15 \textcolor{lb}{(+9.23)} & 31.61 $\pm$ 0.28 \textcolor{lb}{(+7.75)} \\
    Instant-Teaching~\cite{Instant-Teaching_CVPR21} & CVPR21  & 18.05 $\pm$ 0.15 \textcolor{lb}{(+9.00)} & 26.75 $\pm$ 0.05 \textcolor{lb}{(+8.28)} & 30.40 $\pm$ 0.05 \textcolor{lb}{(+6.54)} \\
    Soft Teacher~\cite{SoftTeacher}& ICCV21 & 20.46 $\pm$ 0.39 \textcolor{lb}{(+11.41)} & 30.74 $\pm$ 0.08 \textcolor{lb}{(+12.27)} & 34.04 $\pm$ 0.14 \textcolor{lb}{(+10.18)} \\
    PseCo~\cite{PseCo_ECCV22} & ECCV22 & 22.43 $\pm$ 0.36 \textcolor{lb}{(+13.38)} & 32.50 $\pm$ 0.08 \textcolor{lb}{(+14.03)} & 36.06 $\pm$ 0.24 \textcolor{lb}{(+12.2)} \\
    \midrule
    DINO~\cite{dino_detr_ICLR_23} (Supervised) & - & 18.00 $\pm$ 0.21 & 29.50 $\pm$ 0.16 & 35.00 $\pm$ 0.12 \\
    Omni-DETR~\cite{omni-DETR_cvpr22} (DINO) & CVPR22 & 27.60 \textcolor{lb}{(+9.60)} & 37.70\textcolor{lb}{(+8.20)} & 41.30 \textcolor{lb}{(+6.30)} \\
    Semi-DETR~\cite{Semi-DETR_cvpr23} (DINO) & CVPR23 & 30.5 $\pm$ 0.30 \textcolor{lb}{(+12.50)} & 40.10 $\pm$ 0.15 \textcolor{lb}{(+10.6)} & 43.5 $\pm$ 0.10 \textcolor{lb}{(+8.5)} \\
    \rowcolor{gray!25} \textbf{Sparse Semi-DETR} & - &  \textbf{30.9 $\pm$ 0.23 \textcolor{lb}{(+12.90)}} &\textbf{40.8 $\pm$ 0.12 \textcolor{lb}{(+11.30)}} & \textbf{44.3 $\pm$ 0.01 \textcolor{lb}{(+9.30)}}\\
    \bottomrule
  \end{tabular}
  \end{adjustbox}
 \caption{\textbf{Comparing Sparse Semi-DETR with other approaches on COCO-Partial setting.} The results are the average across all five folds. Under the COCO-partial setting, FCOS serves as the baseline for one-stage detectors, Faster RCNN for two-stage detectors, and DINO for transformer-based end-to-end detectors.}
 \label{tab:results_coco_partial}
\end{minipage}
\end{table*}
\begin{table*}[t]
\hfill
\begin{minipage}[]{0.32\linewidth}
\centering
%\footnotesize
\begin{adjustbox}{width=\linewidth}
 \begin{tabular}{l c >{\columncolor{gray!20}}c c >{\columncolor{gray!20}}c}
 \toprule
    \multirow{2}{*}{Methods} & \multirow{2}{*} {Labels} &
      \multicolumn{3}{c}{COCO-Partial} \\
      % \vspace{2pt}
      \cmidrule(lr){3-5} & & { $AP_{S}$} & {$AP_{M}$} & {$AP_{L}$} \\
   \midrule
   \multirow{3}{*}{Semi-DETR~\cite{Semi-DETR_cvpr23}}  & 1\% & 13.6 & 31.2 & 40.8 \\
                               & 5\% & 23.0  &  43.1 & 53.7 \\
                               & 10\%   & 25.2 &  46.8 & 58.0 \\
    \midrule
    \multirow{3}{*}{Sparse Semi-DETR }& 1\% & 14.8  & 32.5 & 41.4 \\
                                      & 5\% & 23.9 & 44.2 & 54.2 \\
                                      & 10\% & 26.9 & 48.0 & 59.6 \\
    \bottomrule
  \end{tabular}
\end{adjustbox}
\caption{\textbf{Experimental results on COCO-partial settings for small, medium, and large objects}. The results shown are the average across all five folds. We reproduce Semi-DETR results using their source code.}
\label{tab:results_small}
\end{minipage}
%\vspace{5pt}
\hfill
\begin{minipage}[]{0.30\linewidth}
\centering
%\footnotesize
\begin{adjustbox}{width=.99\linewidth}
 \begin{tabular}{l >{}c c }
 \toprule
    \multirow{2}{*}{Methods} &
      \multicolumn{2}{c}{VOC12} \\
      % \vspace{2pt}
      \cmidrule(lr){2-3} & { $AP_{50}$} & {$AP_{50:95}$} \\
   \toprule
   FCOS~\cite{FCOS} (Supervised) & 71.36 & 45.52 \\
   DSL~\cite{DSL_CVPR22} & 80.70  &  56.80 \\
    Dense Teacher~\cite{DenseTeacher} & 79.89 &  55.87\\
    \midrule
    Faster RCNN~\cite{FasterRCNN} (Supervised) & 72.75  & 42.04 \\
    STAC~\cite{STAC}  &  77.45 &  44.64 \\
    HumbleTeacher~\cite{HumbleTeachers_CVPR21} & 80.94 & 53.04  \\
    Instant-Teaching~\cite{Instant-Teaching_CVPR21} & 79.20  & 50.00  \\
    \midrule
    DINO~\cite{dino_detr_ICLR_23} (Supervised)  & 81.20 & 59.60  \\
    Semi-DETR~\cite{Semi-DETR_cvpr23} (DINO) & 86.10 & 65.20 \\
   \rowcolor{gray!20}  \textbf{Sparse Semi-DETR} & \textbf{86.30}  & \textbf{65.51}\\
    \bottomrule
  \end{tabular}
\end{adjustbox}
\caption{\textbf{Experimental results on Pascal VOC protocol}. Here, FCOS, Faster RCNN, and DINO are the supervised baselines.}
\label{tab:results_voc}
\end{minipage}
%\vspace{15pt}
\hfill
\begin{minipage}{0.32\linewidth}
\begin{adjustbox}{width=.98\linewidth}
\centering
  \begin{tabular}{l c} 
    \toprule
    Method & COCO-Full (100\%)\\
    \midrule
    STAC~\cite{STAC}~(18$\times$) & 39.5 $\xrightarrow{\text{-0.3}}$ 39.2 \\
    Unbiased Teacher~(9$\times$) & 40.2 $\xrightarrow{\text{+1.1}}$ 41.3 \\
    SoftTeacher~\cite{SoftTeacher}~(24$\times$) & 40.9 $\xrightarrow{\text{+3.6}}$ 44.5 \\
    DSL~\cite{DSL_CVPR22}~(12$\times$) & 40.2 $\xrightarrow{\text{+3.6}}$ 43.8 \\
    Dense Teacher~\cite{DenseTeacher}~(18$\times$) & 41.2 $\xrightarrow{\text{+3.6}}$ 46.1 \\
    PseCo~(24$\times$) & 41.0 $\xrightarrow{\text{+5.1}}$ 46.1 \\
    Instant-Teaching~\cite{Instant-Teaching_CVPR21}~($24\times$) & 37.6 $\xrightarrow{\text{-0.27}}$ 40.2 \\
    Semi-DETR~\cite{Semi-DETR_cvpr23}~(8$\times$) & 48.6 $\xrightarrow{\text{+1.8}}$ 50.4 \\
     \rowcolor{gray!25} \textbf{Sparse Semi-DETR}~(8$\times$) &  \textbf{49.2 $\xrightarrow{\text{+2.1}}$ 51.3 } \\
\bottomrule
 \end{tabular}  
 \end{adjustbox}
 \caption{\textbf{Comparing Sparse Semi-DETR with other approaches on COCO-Full}. Note that 1 denotes 30k training iterations, while an N$\times$ signifies N times 30k iterations.}
 \label{tab:results_coco_full}
\end{minipage}
\end{table*}

\noindent\textbf{COCO-Partial benchmark.} 
Sparse Semi-DETR outperforms the current SSOD methods in COCO-Partial across all experiment settings, as demonstrated in Table~\ref{tab:results_coco_partial}. 
(1) We compare our method to both one-stage and two-stage SSOD. Sparse Semi-DETR surpasses Dense Teacher by 8.52, 7.79, 7.17 mAP on 1\%, 5\%, and 10\% label data. It also outperforms PseCo by 8.47, 8.30, 8.24 mAP on 1\%, 5\%, and 10\% label data. Sparse Semi-DETR's superior performance as a semi-supervised object detector is achieved without needing hand-crafted components commonly used in two-stage and one-stage detectors. (2) When compared to DETR-based detectors, Sparse Semi-DETR outperforms omni-DETR by 3.30, 3.10, and 3.00 mAP and beats Semi-DETR by 0.40, 0.70, 0.80 mAP on 1\%, 5\%, and 10\% label data. (3) Sparse Semi-DETR's exceptional proficiency in precisely detecting small and partially obscured objects is a standout feature. In Figure~\ref{fig:result1}, we visually compare Sparse Semi-DETR with the two preceding approaches using the COCO 10\% labeled dataset. These results demonstrate the impressive capabilities of Sparse Semi-DETR, particularly in its ability to identify small objects and objects concealed by obstacles, as highlighted in the third-row images by the white arrows. Table~\ref{tab:results_small} exhibits a remarkable performance boost of Sparse Semi-DETR on small objects. It surpasses the Semi-DETR by 1.20, 0.90 and 1.70 mAP on small objects using 1\%, 5\%, and 10\% label data, respectively. It highlights the superior efficiency and accuracy of Sparse Semi-DETR in detecting smaller objects.

\noindent\textbf{COCO-Full benchmark.} 
In Table~\ref{tab:results_coco_full}, when incorporating additional \texttt{unlabel2017} data, Sparse Semi-DETR demonstrates a substantial improvement, achieving an impressive 2.1 mAP gain and reaching a total of 51.3 mAP. It surpasses the performance of Dense Teacher, PseCo, and Semi-DETR by 5.2, 5.2, and 0.9 mAP, respectively, highlighting the effectiveness of Sparse Semi-DETR. 

\noindent\textbf{Pascal VOC benchmark.} 
Sparse Semi-DETR exhibits a remarkable performance boost on the Pascal VOC benchmark, as shown in Table~\ref{tab:results_voc}. It surpasses the supervised baseline by 5.1 improvement on $AP_{50}$ and by 5.91 impressive increase on $AP_{50:95}$. Furthermore, it outperforms all previously single-stage, two-stage, and DETR-based SSOD methods by a significant margin.

\begin{table*}[ht]
\centering
\begin{subtable}[ht]{0.50\linewidth}
\begin{adjustbox}{width=\linewidth}
\begin{tabular}{c|c|c|c|c}
    \toprule
    Query Refinement & Pseudo-Label Filtering & mAP & $AP_{50}$ & $AP_{75}$ \\
    \midrule
    \color{black}\xmark  & \color{black}\xmark & 43.5 & 58.9 & 46.0 \\
    \color{black}\checkmark & \color{black}\xmark & 43.8 & 61.1 & 47.3\\
    \color{black}\xmark & \color{black}\checkmark & 43.9 & 60.5 & 46.3\\
    \rowcolor{gray!20} \color{black}\checkmark & \color{black}\checkmark &  44.3 & 61.7 & 47.6 \\
    \bottomrule
\end{tabular}
\end{adjustbox}
\centering
\caption{\textbf{Effect of Individual Component.}}
\label{tab:effect_each_module_a}
\end{subtable} 
\begin{subtable}[ht]{0.33\linewidth}
\begin{adjustbox}{width=.95\linewidth}
\begin{tabular}{c|c|c|c|c}
    \toprule
     Teacher &  Student&  mAP & $AP_{50}$ & $AP_{75}$ \\
    \midrule
     \color{black}\xmark  & \color{black}\xmark  &  43.5 & 58.9 & 46.0 \\
    \color{black}\checkmark  & \color{black}\xmark  & 43.8  & 61.2 & 47.2 \\
   \rowcolor{gray!20} \color{black}\xmark &  \color{black}\checkmark & 44.3 & 61.7 & 47.6 \\
   \color{black}\checkmark  & \color{black}\checkmark  &  42.8 & 59.7 & 45.8\\
    \bottomrule
\end{tabular}
\end{adjustbox}
\centering
\caption{\textbf{Effect of QR on Student and Teacher module.}}
\label{tab:ST_attention_b}
\end{subtable}
\begin{subtable}[ht]{0.24\linewidth}
\begin{adjustbox}{width=.98\linewidth}
\begin{tabular}{c|c|c|c}
    \toprule
     MLP &  mAP & $AP_{50}$ & $AP_{75}$ \\
    \midrule
    \checkmark & 43.7 &  60.9 & 46.9  \\
     \xmark & \cellcolor{gray!20} 44.3 & \cellcolor{gray!20}61.7 & \cellcolor{gray!20} 47.6\\
    \bottomrule
\end{tabular}
\end{adjustbox}
\centering
\caption{\textbf{Effect of MLP.}}
\label{tab:ROI_f}
\end{subtable}
\begin{subtable}[ht]{0.35\linewidth}
\begin{adjustbox}{width=.95\linewidth}
\begin{tabular}{c|c|c|c}
    \toprule
    Method &  mAP & $AP_{50}$ & $AP_{75}$ \\
    \midrule
     Single-view Queries & 43.0 &  59.3 & 46.3 \\
    Cross-view Queries & 43.5 &  58.9 & 46.0  \\
   \rowcolor{gray!20} Query Refinement &  44.3 & 61.7 & 47.6\\
    \bottomrule
\end{tabular}
\end{adjustbox}
\centering
\caption{\textbf{Effect of different variants of queries.}}
\label{tab:quries_variants_g}
\end{subtable}
\begin{subtable}[ht]{0.35\linewidth}
\begin{adjustbox}{width=.92\linewidth}
\begin{tabular}{c|c|c|c}
\toprule
    Method  & mAP & $AP_{50}$ & $AP_{75}$ \\
    \midrule
    Simple Concat & 43.4 & 58.8 & 46.1 \\
    Cosine Similarity & 43.7 & 60.3 & 46.1 \\
    \rowcolor{gray!20} Attention Module & 44.3 & 61.7 & 47.6 \\
    \bottomrule
\end{tabular}
\end{adjustbox}
\centering
\caption{\textbf{Effectiveness of Attentional module in QR.}}
\label{tab:attention_queries_h}
\end{subtable}
\caption{\textbf{Ablations for the proposed Sparse Semi-DETR on COCO 10\% Label dataset.} \textbf{(a)} We analyze the effectiveness of each module of Sparse Semi-DETR. \textbf{(b)} We experiment with different QR combinations to identify the optimal design, applying QR selectively on the student, the teacher, or both. \textbf{(c)}  We analyze the effect of using MLP layers for $F_{t1}$ and $F_{s1}$. Empirical observation reveals that we do not require MLP as in~\cite{Semi-DETR_cvpr23}. \textbf{(d)} We observe the effect of the attentional block in the Query Refinement module by replacing it with simple concatenation or cosine similarity. \textbf{(e)} We vary the type of object queries fed to the decoder with the decoder's original queries. The best results are \colorbox{gray!20}{highlighted}.}
\label{tab:AS_queries}
\vspace{-1pt}
\end{table*}
\begin{figure}
\centering
\includegraphics[width=\linewidth]{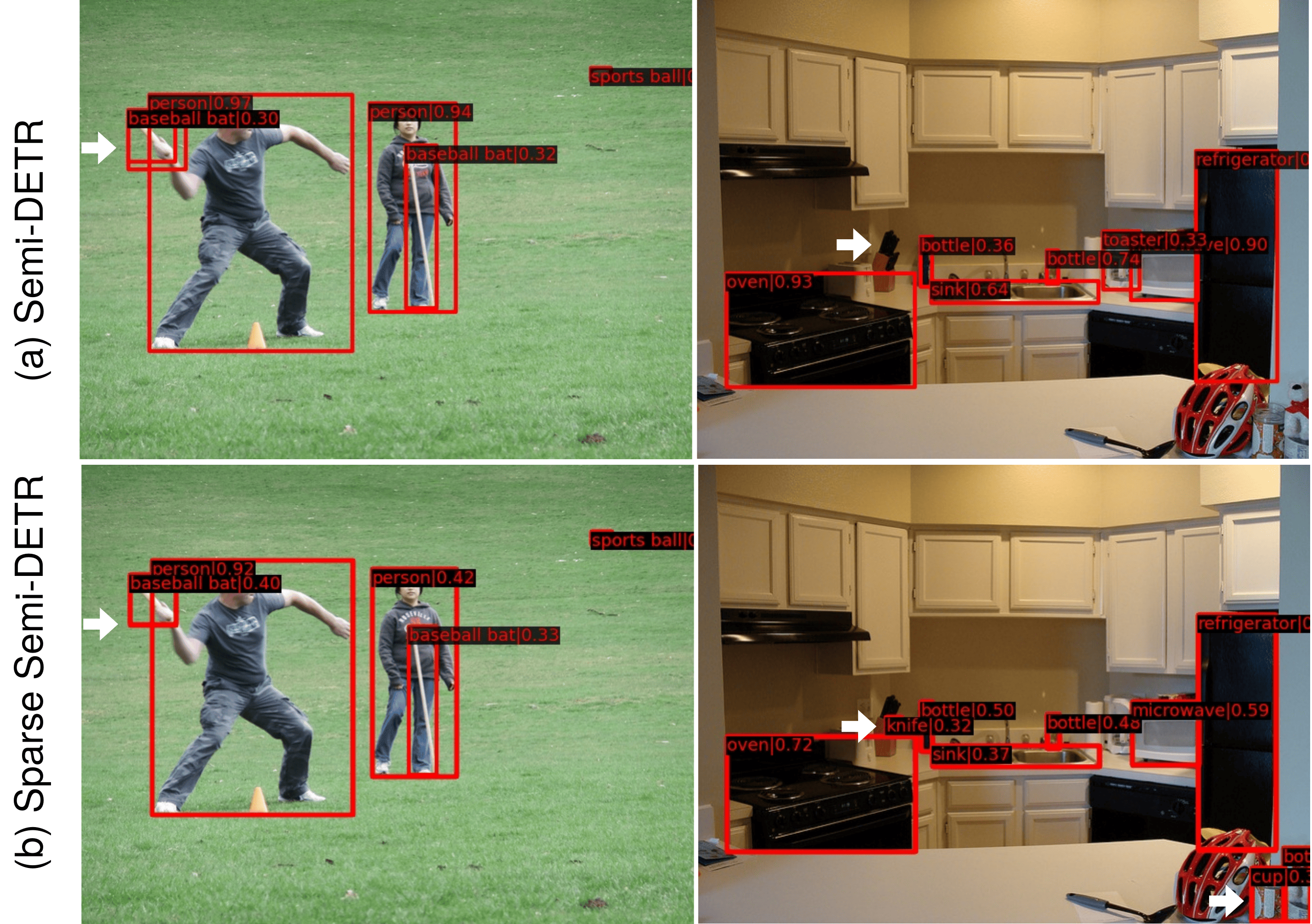}
   \caption{
   %The top-row figures show the detection results of Semi-DETR, while the bottom-row figures show the results using Sparse Semi-DETR. Both networks trained for 120k iterations using the one-to-many and one-to-one training strategies. In the bottom-left image, Sparse Semi-DETR removes redundant bounding boxes indicated by a white arrow. In the top-right image, small objects like knives, indicated by a white arrow, are detected by Sparse-Semi DETR.
   The top-row figures display Semi-DETR's detection results, and the bottom-row shows Sparse Semi-DETR's outcomes. Both networks were trained for 120k iterations using one-to-many and one-to-one strategies. Sparse Semi-DETR eliminates redundant bounding boxes in the bottom-left image and detects small objects, like knives, in the top-right image, as indicated by white arrows.
   }
\label{fig:result2}
%\vspace{-10pt}
\end{figure} 
\subsection{Ablation Studies}
\label{sec:ablation_studies} 
This section ablates the key design choices of Sparse Semi-DETR. The experiments detailed in this section are executed on the MS COCO dataset with 10\% label data, employing DINO as the primary detector.

\noindent\textbf{Effect of Individual Component.}
We conduct three experiments to assess the efficacy of each module of Sparse Semi-DETR, as detailed in Table~\ref{tab:effect_each_module_a}.~We employ Semi-DETR as our baseline model. Notably, integrating each component into Sparse Semi-DETR leads to consistent performance improvements. Specifically, adding the Query Refinement and Pseudo-Label Filtering modules yields a significant increase of 0.8 mAP. These results confirm that each component of Sparse Semi-DETR enhances our model's performance. 

\begin{table}[ht]
\centering
\begin{subtable}{0.30\linewidth}
\begin{adjustbox}{width=\linewidth}
\begin{tabular}{c|c}
    \toprule
    m (k=4) & mAP   \\
    \midrule
    2 &  44.0 \\
    4 &  44.1 \\
    \rowcolor{gray!20} 6 & 44.3  \\
    8 & 44.2 \\
    \bottomrule
\end{tabular}
\end{adjustbox}
\centering
\caption{\textbf{Effect of m.}}
\label{tab:LR_m}
\end{subtable}
\hspace{1pt}
\begin{subtable}{0.30\linewidth}
\begin{adjustbox}{width=\linewidth}
\begin{tabular}{c|c}
    \toprule
    k (m=6) & mAP   \\
    \midrule
    2 & 43.8 \\
    \rowcolor{gray!20} 4 & 44.3  \\
    6 & 44.1 \\
    8 & 43.9 \\
    \bottomrule
\end{tabular}
\end{adjustbox}
\centering
\caption{\textbf{Effect of k.}}
\label{tab:LR_k}
\end{subtable}
\hspace{1pt}
\begin{subtable}{0.23\linewidth}
\begin{adjustbox}{width=.98\linewidth}
\begin{tabular}{c|c}
    \toprule
    $\sigma$ & mAP   \\
    \midrule
    0.2 & 43.2 \\
    0.3 & 43.8 \\
    \rowcolor{gray!20} 0.4 & 44.3  \\
    0.5 & 44.0 \\
    \bottomrule
\end{tabular}
\end{adjustbox}
\centering
\caption{\textbf{Effect of $\sigma$.}}
\label{tab:LR_a}
\end{subtable}
\caption{\textbf{Ablations for the Reliable Pseudo-Label Filtering Module on COCO 10\% Label dataset.} We study the impact of various parameters as augmented ground truth (m), top pseudo-labels selection (k) and filtering threshold ($\sigma$).}
\label{tab:AS_label}
\end{table}

\noindent\textbf{Effect of Query Refinement Module.}
We examine the impact of the Query Refinement (QR) Module. In Table~\ref{tab:ST_attention_b}, we explore various QR combinations to determine the most effective design. Applying QR selectively to the student, teacher, or both networks, we find that employing QR on weak augmented image features $F_{t1}$ and $F_{t2}$, and integrating them into the student network with the original decoder queries yields the best results. Table~\ref{tab:ROI_f} shows that processing $F_{t1}$ features without MLP (in~\cite{Semi-DETR_cvpr23}) improves results. Table~\ref{tab:quries_variants_g} presents our study on the impact of different query variants, where we observe that QR consistently outperforms other methods. Finally, Table~\ref{tab:attention_queries_h} shows that using an attentional block in the QR module is more effective than simple concatenation or cosine similarity. We provide more analysis of the Query Refinement Module supplementary document. Figure~\ref{fig:result2} compares Sparse Semi-DETR and Semi-DETR. Sparse Semi-DETR, processing fewer but refined queries, demonstrates lower duplication rates in this training approach.
\begin{table}[h]
\centering
\begin{adjustbox}{width=.75\linewidth}
\begin{tabular}{cccccccc}
    \toprule
     Iterations & 40k & 60k & 80k & 100k & 120k  \\
    \midrule
     mAP & 43.5 &  \cellcolor{gray!20}44.3 & 44.0 & 43.8 & 44.6\\
       without-NMS & Y & Y & Y & Y & N \\
    \bottomrule
\end{tabular}
\end{adjustbox}
\centering
\caption{Evaluating One-to-Many Assignment Strategy.}
\label{tab:NMS}
\end{table}

\noindent\textbf{Effect of Reliable Pseudo-Label Filtering Module.} 
In our analysis of the Pseudo-Label Filtering Module, we examine the impact of various parameters. Table~\ref{tab:LR_m} shows that a smaller m results in lower performance due to the inclusion of poor-quality labels. Performance improves with a moderately larger m due to enhanced auxiliary loss. However, excessively large m values trigger NMS, negatively impacting performance. Additionally, in Table~\ref{tab:LR_k}, we analyze the selection of the k value and find that setting k to 4 yields the best performance. In Table~\ref{tab:LR_a}, the optimal performance for $\sigma$ is achieved at 0.4; values lower than this may lead to the generation of noisy pseudo-labels, whereas higher values can decrease the number of effective pseudo-labels.

\noindent\textbf{Limitation.} We still have duplications when we apply the one-to-many training strategy in both stages. Table~\ref{tab:NMS} illustrates that a one-to-many strategy for 120k iterations with Sparse Semi-DETR achieves a 44.6 mAP using NMS. In comparison, 60k iterations in the first stage attain a comparable 44.3 mAP without NMS. Future works can explore this aspect.%We leave this for future works to explore.

\section{Conclusion}
\label{sec:conclusion}
In conclusion, we successfully address the inherent limitations of DETR-based semi-supervised object detection frameworks by introducing Sparse Semi-DETR. This novel solution effectively tackles overlapping predictions and the detection of small objects. Sparse Semi-DETR incorporates a Query Refinement Module to enhance object query quality, mainly benefiting the detection of small and partially obscured objects. Besides, it also introduces a Reliable Pseudo-Label Filtering Module to filter out low-quality pseudo-labels selectively, thereby enhancing overall detection accuracy and consistency with the remaining high-quality labels. Our method outperforms existing SSOD approaches, with extensive experiments demonstrating its effectiveness.
%Extensive experiments on the MS COCO and Pascal VOC benchmarks demonstrate the effectiveness of our method.
%This is accomplished by improving the quality of object queries and filtering out the noisy pseudo labels, crucial aspects often overlooked in traditional DETR models.

\noindent\textbf{Ethical considerations.} We study semi-supervised models, and agree that standard ethical considerations for visual recognition are applicable to our work.
\section*{Acknowledgements}
This work was in parts supported by the EU Horizon Europe Framework under grant agreements 101135724 (LUMINOUS) and 101092312 (AIRISE) and by the Federal Ministry of Education and Research of Germany (BMBF) under grant 01QE2227C (HERON).

% This work received partial support from the European Union's Horizon Europe Framework Programme under grant agreements 101135724 (LUMINOUS) and 101092312 (AIRISE). Additionally, it was partly funded by the Federal Ministry of Education and Research of the Federal Republic of Germany (BMBF) under grant agreement 01QE2227C (HERON).


\thispagestyle{empty}
{
    \small
    \bibliographystyle{ieeenat_fullname}
    \bibliography{main}
}

%\maketitlesupplementary
\clearpage
\maketitlesupplementary
%\section*{Supplementary Material}
The supplementary document offers an extensive overview of our approach, detailed insights into our implementation details, and a comprehensive analysis of results.
\section*{A1.1 Additional Details of Sparse Semi-DETR} 
The Sparse Semi-DETR framework is an extension of Semi-DETR (the first semi-supervised DETR-based framework). Labeled data is used for student network training, employing a supervised loss. The Sparse Semi-DETR framework processes unlabeled data through two distinct pathways: the teacher network, which receives weakly augmented data, and the student network, which is fed with strongly augmented data. The teacher network utilizes the unlabeled data to produce pseudo-labels. Meanwhile, the student model undergoes parameter refinement via back-propagation. In contrast, the teacher model's parameters are updated, following the exponential moving average (EMA) of the student model. 
\begin{table*}[h!]
\begin{center}
\begin{tabular}{ c|c|c|c|c} 
\hline
training setting & COCO-Partial & COCO-Full & VOC & Ablation  \\
\hline
batch size & 5*8 & 8*8 & 5*8 & 5*8  \\
labeled to unlabeled data ratio & 1:4 & 1:1 & 1:4 & 1:4  \\
learning rate & 0.001 & 0.001 & 0.001 & 0.001  \\
first stage iterations & 0-60K & 0-180K & 0-40K & 0-60K  \\
second stage iterations & 60k-120K & 180k-240K & 40k-60K & 60k-120K  \\
iterations & 120K & 240K & 60K & 120K  \\
unsupervised loss weight \( \alpha \) & 4.0 & 2.0 & 4.0 & 4.0  \\
EMA rate & 0.996 & 0.999 & 0.999 & 0.999  \\
confidence threshold & 0.4 & 0.4 & 0.4 & 0.4  \\
\hline
\end{tabular}
\caption{Training settings for different datasets. Here, `Ablation' means the training setting of the ablation studies in the paper.}
\label{tab:setting}
\end{center}
\end{table*}

\noindent\textbf{Additional Details of Semi-DETR.}
Semi-DETR is a DETR-based semi-supervised framework that introduces cross-view query consistency and stage-wise hybrid matching strategies. (1) In CNN-based semi-supervised object detection (SSOD) frameworks~\cite{SoftTeacher,Mean_Teachers_CVPR21,Instant-Teaching_CVPR21,Consistent_Teacher_CVPR23,DSL_CVPR22,SED_CVPR22,CSL_NIPS19,PseCo_ECCV22,UnbiasedTeacherv2_CVPR22}, consistency regularization is easily implemented by minimizing differences between teacher and student model outputs, given the same input but with different augmentations. However, this approach is not directly applicable in DETR-based SSOD frameworks due to the lack of clear correspondence between input object queries and output predictions. To address this, a novel cross-view query consistency module is proposed. 
%This module enables DETR-based models~\cite{deformable_detr,UPDETR_CVPR20,smca23,CondDE,WBdetr4,pnp6,yolos6,fpdetr,rego2} to learn semantically invariant characteristics of object queries across different augmented views.
It processes RoI features through MLPs, and generates cross-view query embeddings. These embeddings are combined with original object queries and fed into a decoder. (2) Semi-DETR initially uses a one-to-many assignment in early training, allowing multiple predictions per pseudo-label. It speeds up convergence and improves label quality but can cause redundant predictions. It then switches to one-to-one assignment, reducing redundancy and aiming for an NMS-free final model. However, its effectiveness on small objects is limited. Our Sparse semi-DETR refines object queries, enhancing small object detection and accuracy.
\section*{A1.2 Additional Details of Implementation.}
The implementation of the Sparse Semi-DETR approach is based on MMdetection  framework~\cite{mmdetection}. 
%During our training process, we employ the Focal Loss~\cite{retinaNet} for the classification task and Smooth L1 Loss and GIoU Loss~\cite{GIOU2} for regression.
We integrate data pre-processing methodologies from Soft-Teacher~\cite{SoftTeacher}. We train the network on 8 GPUs (RTXA6000), which takes roughly two training days to complete 120k training iterations. Elaborating on training hyperparameters for different benchmarks: (1) COCO-Partial Setup:~We train the network using 8 GPUs for 120k iterations, with each GPU handling five images. It employs one-to-many assignment strategy for first 60k iterations and then one-to-one assignment strategy for 60k-120k iterations. 
%We train for 120k iterations on 8 GPUs, with five images per GPU. 
%The ratio of labeled to unlabeled data is set to 1:4. 
(2) COCO-Full Setup:~For this benchmark, we train for 240k iterations, employing one-to-many assignment strategy for first 180k iterations and then one-to-one assignment strategy for 180k-240k iterations. We use 8 GPUs with eight images per GPU. 
%The ratio of labeled to unlabeled data is set to 1:1. 
(3) Pascal VOC Setup:~Here, first 40k iterations adopt a one-to-many assignment strategy and then one-to-one assignment strategy for 40k-60k iterations. Across all our experimental setups, we've kept the confidence threshold constant at 0.4. We use the Adam optimizer and set the learning rate to 0.001. We avoid using learning rate decay for a fair comparison with Semi-DETR~\cite{Semi-DETR_cvpr23}. Complete implementation details are provided in Table~\ref{tab:setting}. 

%In this benchmark, we train Sparse Semi-DETR for 60k iterations, employing one-to-many assignment strategy for first 40k iterations and then one-to-one assignment strategy for 40k-60k iterations. 
\noindent\textbf{Data Augmentation.}
We adopt the same data augmentation scheme as in Semi-DETR, detailed in Table~\ref{tab:augmentation}. We employ weak augmentation on unlabeled data for generating pseudo labels, while strong augmentation is utilized for both labeled and unlabeled data during the model's training. 

\begin{table*}
\begin{adjustbox}{width=.95\linewidth}
\begin{tabular}{ c|c|c|c } 
\hline
Augmentation & Labeled image training & Unlabeled image training & Pseudo-label generation \\
\hline
Scale Jitter & shortest edge \(\in [480, 800]\) & shortest edge \(\in [480, 800]\) & shortest edge \(\in [480, 800]\) \\
\hline
Solarize Jitter & \(p=0.25\), ratio\(\in(0,1)\) & \(p=0.25\), ratio\(\in(0,1)\) & - \\
\hline
Brightness & \(p=0.25\), ratio\(\in(0,1)\) & \(p=0.25\), ratio\(\in(0,1)\) & - \\
\hline
Contrast Jitter & \(p=0.25\), ratio\(\in(0,1)\) & \(p=0.25\), ratio\(\in(0,1)\) & - \\
\hline
Sharpness Jitter & \(p=0.25\), ratio\(\in(0,1)\) & \(p=0.25\), ratio\(\in(0,1)\) & - \\
\hline
Translation & - & \(p=0.3\), translation ratio\(\in(0,1)\) & - \\
\hline
Rotate & - & \(p=0.3\), angle\(\in(0,30^\circ)\) & - \\
\hline
Shift & - & \(p=0.3\), angle\(\in(0,30^\circ)\) & - \\
\hline
Cutout & num\(\in(1,5)\), ratio\(\in(0.05,0.2)\) & num\(\in(1,5)\), ratio\(\in(0.05,0.2)\) & - \\
\hline
\end{tabular}
\end{adjustbox}
\centering
\caption{Data augmentations used in our approach. \( p \) indicate the probability of choosing a certain type of augmentation.}
\label{tab:augmentation}
\end{table*}
\section*{A1.3 Additional Details of Results.}
\noindent\textbf{Additional Details of Query Refinement Module.}
We perform additional experiments to assess the efficacy of our query refinement approach as follows:
\begin{enumerate}
    \item Is the attention module crucial in query refinement? Could we apply attention to just low or high-resolution features exclusively, or should it be applied to both high and low-resolution features for optimal results?
    \item Is the integration of a similarity module crucial in query refinement? How would training be impacted if we disregarded similarity features and considered all features comprehensively?
\end{enumerate}
\begin{figure*}
\centering
\includegraphics[width=.70\linewidth]{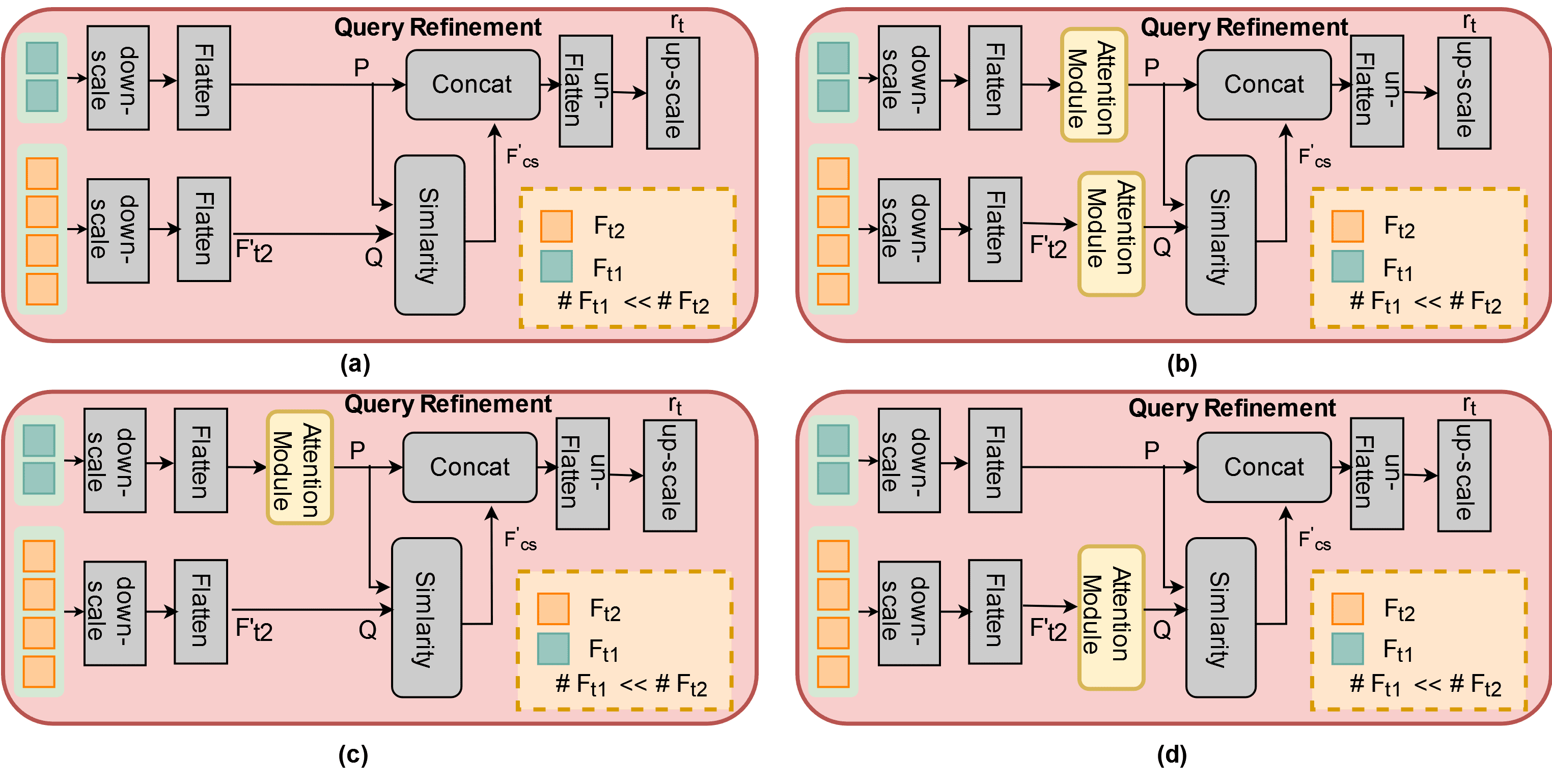}
   \caption{Overview of the Impact of Attention Module in Query Refinement. (a) Query refinement without an attention module, (b) Attention module applied to both low and high-resolution features, (c) Attention module applied to high-resolution features, and (d) Attention module applied to low-resolution features. The best results are achieved for refining queries by applying the attention module to low-resolution features and then combining these with high-resolution features.}
\label{fig:queries-supp1}
\end{figure*} 
\begin{figure*}
\centering
\includegraphics[width=.80\linewidth]{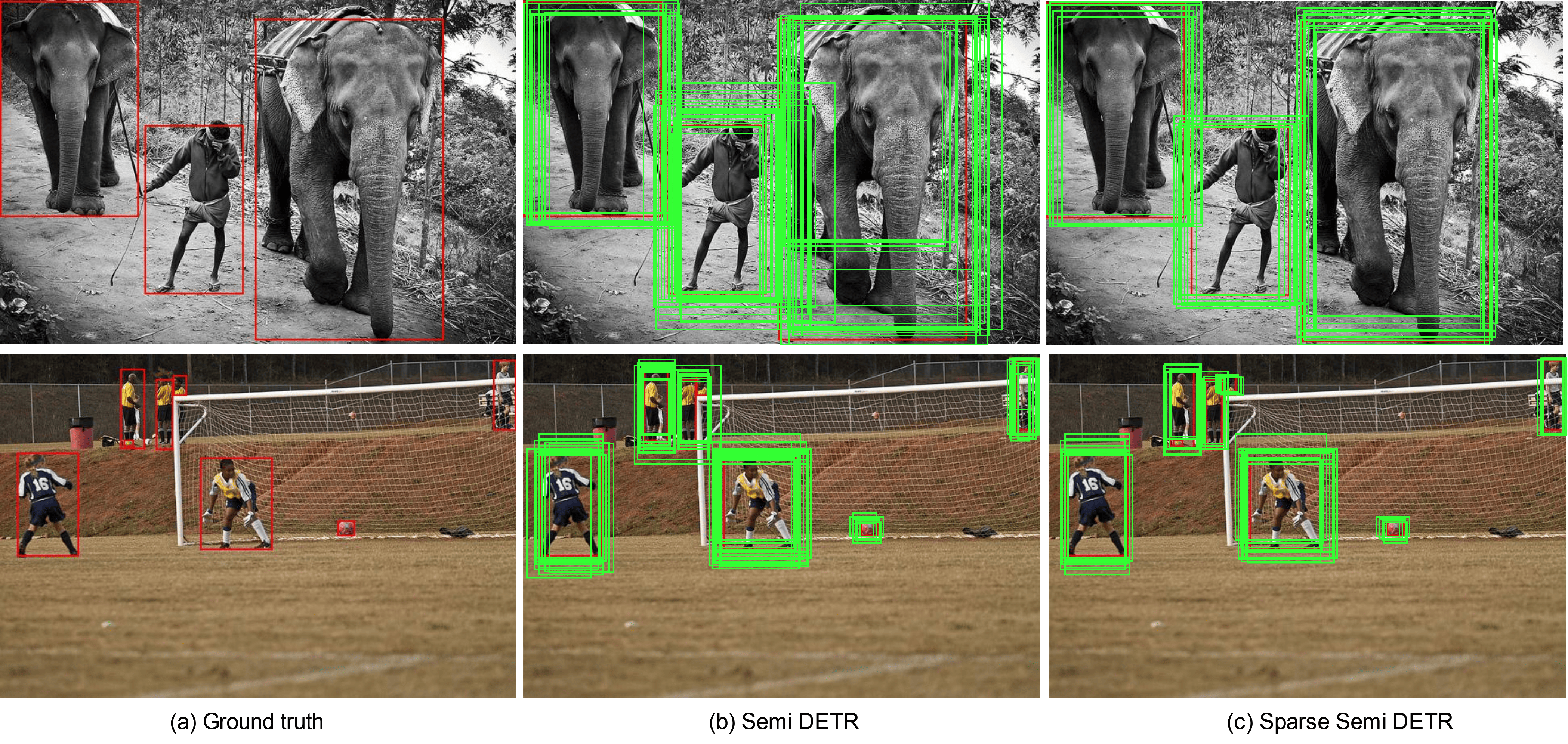}
   \caption{Qualitative Comparison of positive proposals in One-to-Many assignment strategy: (a) Ground Truth (b) Semi-DETR (c) Sparse Semi-DETR. Our approach, compared to Semi-DETR, generates more refined positive proposals for each ground truth. Here, ground truths are outlined in red, while the positive proposals are highlighted in green. Sparse Semi-DETR performs better in identifying small or hidden objects, as indicated by positive proposals around such items. It employs an attention mechanism, focusing on finer image details, which enhances the detection of hidden objects. Additionally, its similarity module further refines the proposal quality, leading to a notably improved identification accuracy.}
\label{fig:queries-results}
\end{figure*} 
\noindent\textbf{Impact of Attention module:}  In our Query Refinement, we refine the queries by applying the attention module on \(F_{t2}\) features and combining them with \(F_{t1}\) features. In this experiment, we study the impact of the attention module in Query Refinement, as highlighted in Table~\ref{tab:attention_network}. Figure~\ref{fig:queries-supp1} (a) illustrates the concatenating high-resolution features after extracting similar features from the low-resolution without applying an attention network to either set of features. Secondly, as indicated in Figure~\ref{fig:queries-supp1} (b), we apply the attention network on both sets of features. In Figure~\ref{fig:queries-supp1} (c), we extract similar features from the \(F_{t2}\) and apply the attention network on just \(F_{t1}\) features. In Figure~\ref{fig:queries-supp1} (d), we apply the attention network on \(F_{t2}\) features and find similarity with \(F_{t1}\) features that gives the best results. The model can focus on capturing essential information by applying the attention module to \(F_{t2}\) features. When these enhanced \(F_{t2}\) features are compared for similarity with \(F_{t1}\) features, the model can get refined detail. It enables the model to make more accurate predictions, leading to better overall performance.
\begin{table}[ht]
\centering
\begin{adjustbox}{width=.75\linewidth}
\begin{tabular}{cccccc}
\toprule
    \multirow{2}{*}{ID} & \multicolumn{2}{c}{Attention}  & \multirow{2}{*}{mAP} & \multirow{2}{*}{$AP_{50}$} & \multirow{2}{*}{$AP_{75}$} \\
      \cmidrule(lr){2-3} & $F_{t1}$ & $F_{t2}$ \\
    \hline
    (a) & \color{black}\xmark   & \color{black}\xmark & 43.4 & 58.8 & 46.1  \\
    (b) & \color{black}\checkmark & \color{black}\checkmark & 40.1 & 57.8 & 43.8 \\
    (c) & \color{black}\checkmark &  \color{black}\xmark  & 42.8 & 59.7 & 45.8\\
    \rowcolor{gray!20} (d) &  \color{black}\xmark  & \color{black}\checkmark & 44.3 & 61.7 & 47.6 \\
    \bottomrule
\end{tabular}
\end{adjustbox}
\centering
\caption{\textbf{Impact of Attention module.} Here, $F_{t1}$ and $F_{t2}$ are the high resolution and low resolution features, respectively. }
\label{tab:attention_network}
%\vspace{-2pt}
\end{table}

\noindent\textbf{Impact of Similarity module:} We reduce the number of queries in query refinement by filtering similar query features in low-resolution features. As indicated in Table~\ref{tab:similarity_network}, removing the similarity module results in a performance decline of 0.3 mAP, increasing the number of queries in a one-to-many training strategy. It confirms the importance of the similarity module in our query refinement strategy. The effectiveness of refined queries using the similarity module is because when enhanced low-resolution features are compared for similarity with high-resolution features, the model can effectively correlate the relevant information from both levels of detail, improving performance.

\begin{table}[ht]
\centering
\begin{adjustbox}{width=.75\linewidth}
\begin{tabular}{cccccc}
\toprule
     \multicolumn{2}{c}{Similarity}  & \multirow{2}{*}{$\#$ Queries ($r_t$)} & \multirow{2}{*}{mAP} & \multirow{2}{*}{$AP_{50}$} & \multirow{2}{*}{$AP_{75}$} \\
      \cmidrule(lr){1-2} $F_{t1}$ & $F_{t2}$ & $\#F_{t1}<<\#F_{t2}$ \\
    \hline
    \color{black}\xmark   & \color{black}\xmark & \#  $(F_{t1}+F_{t2})$& 44.0 & 61.1 & 47.5  \\
    \rowcolor{gray!20} \color{black}\xmark  & \color{black}\checkmark & \# $(2*F_{t1})$  & 44.3 & 61.7 & 47.6 \\
    \bottomrule
\end{tabular}
\end{adjustbox}
\centering
\caption{\textbf{Impact of Similarity module.} Here, $F_{t1}$ and $F_{t2}$ are the high resolution and low resolution features, respectively.}
\label{tab:similarity_network}
\vspace{-15pt}
\end{table}
\begin{figure*}
\centering
\includegraphics[width=.65\linewidth]{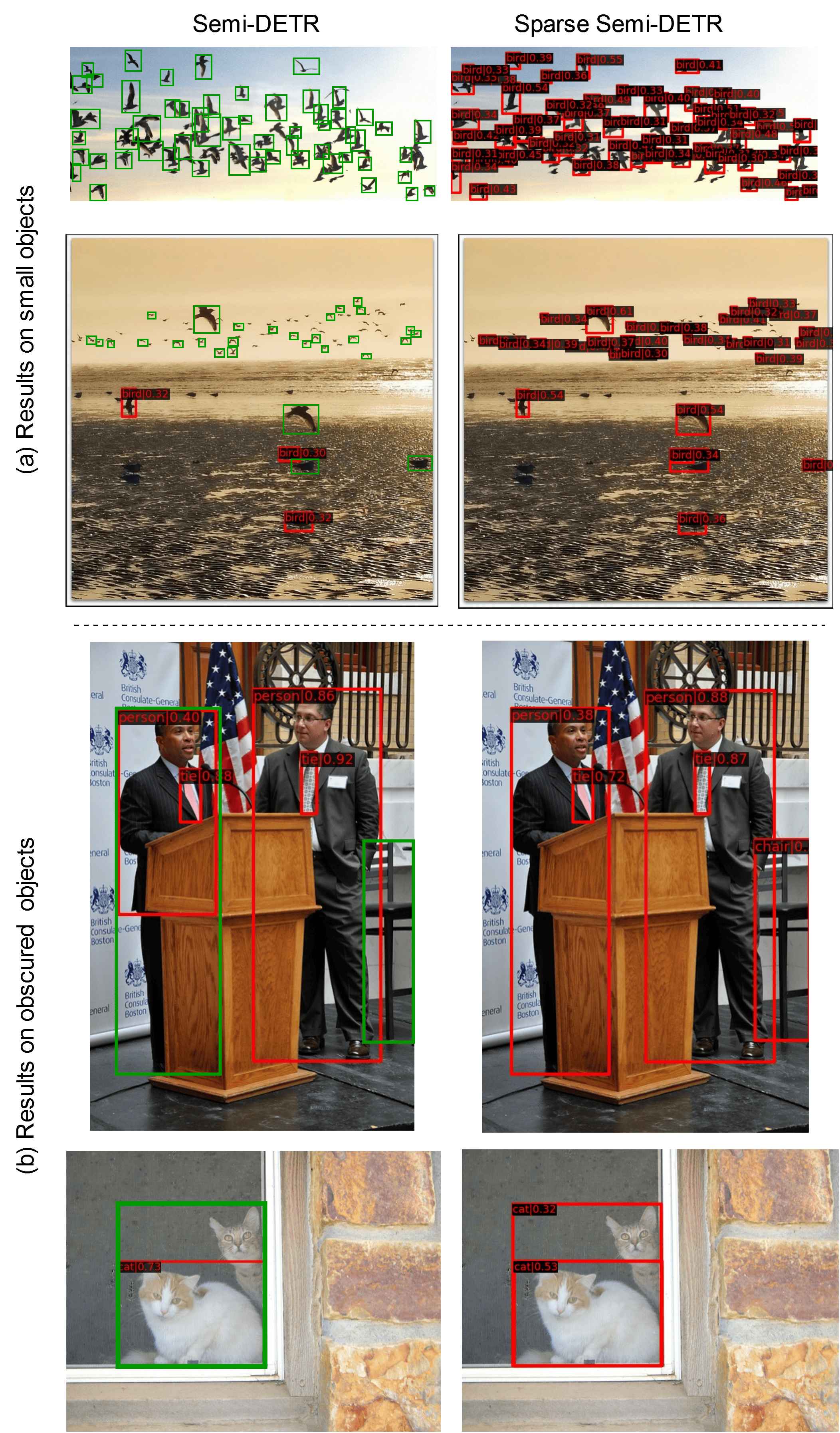}
   \caption{Qualitative comparison on the COCO test set. The prediction results are in red, and the green boxes refer to the prediction difference in Semi-DETR and Sparse Semi-DETR. \textbf{(a) Small Objects:} Semi-DETR, on the left, has missed detections of bird objects, indicated with green bounding boxes as false negatives. On the right, red bounding boxes signify correctly identified birds, showcasing Sparse Semi-DETR's more precise and reliable detection capabilities for smaller objects. \textbf{(b) Obscured Objects:} The green boxes indicate the regions where the Semi-DETR has either failed to detect an object (false negatives) as the chair or incorrectly estimated the region of the objects, like the person. Sparse Semi-DETR detects obscured objects more precisely, improving performance in complex visual environments.}
\label{fig:result6}
\end{figure*} 
\begin{figure*}
\centering
\includegraphics[width=.65\linewidth]{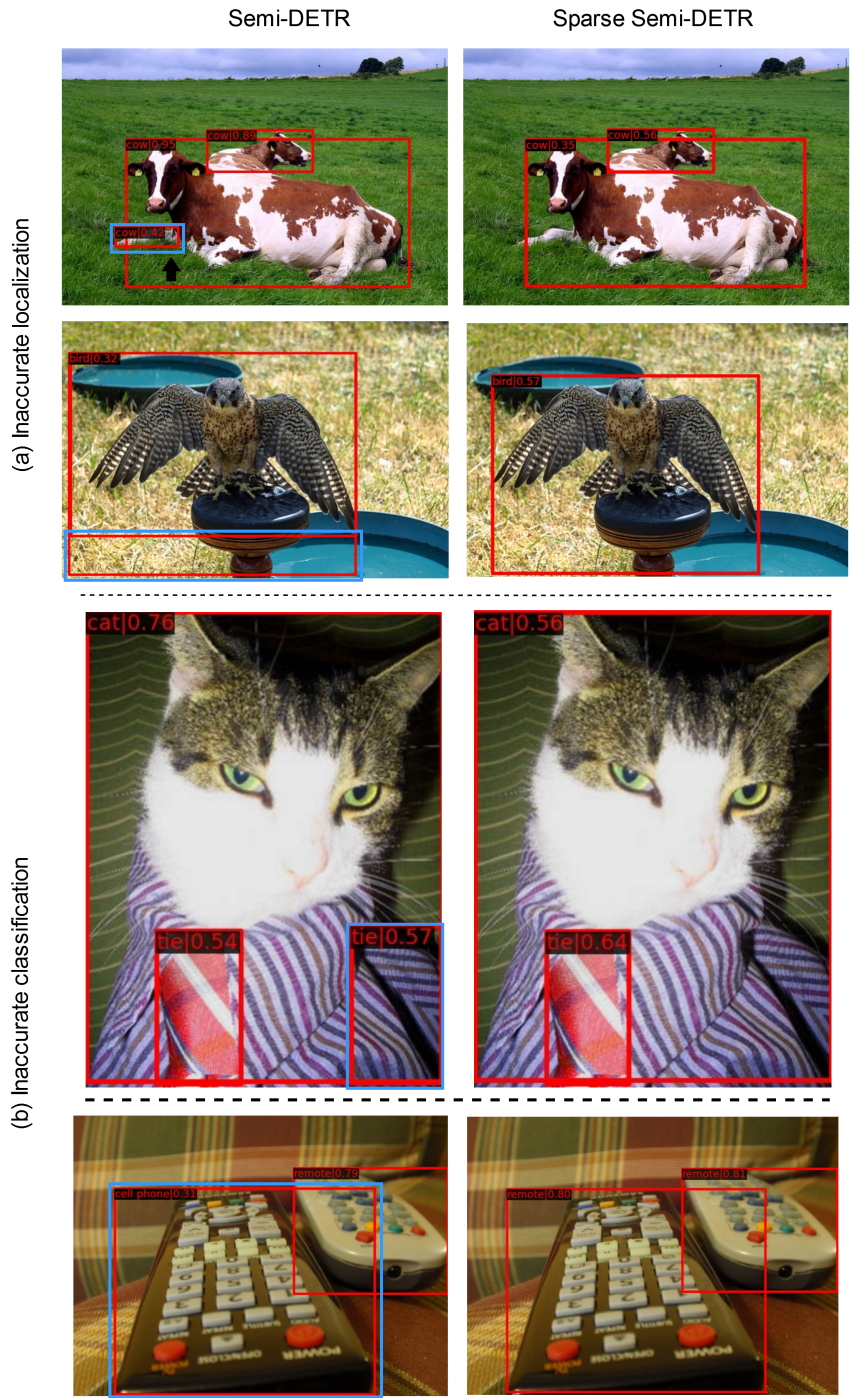}
   \caption{Qualitative comparison on the COCO test data. The prediction results are in red, and the blue boxes highlight the prediction difference in Semi-DETR and Sparse Semi-DETR. \textbf{(a) Inaccurate localization:} Semi-DETR incorrectly places multiple bounding boxes around the individual cow and bird objects, indicating it misidentified them as several entities instead of one. Sparse Semi-DETR, however, shows reduced duplications in bounding boxes. \textbf{(b) Inaccurate classification:} Semi-DETR partially misidentified a cat wearing a tie as the tie itself and confused a 'remote' object with a 'cell phone' as represented with a blue bounding box. Sparse Semi-DETR reduces the misidentification issues present in Semi-DETR, such as not confusing a cat wearing a tie as the tie itself and correctly identifying a 'remote' without mistaking it as a 'cell phone.'}
\label{fig:result7}
\end{figure*} 
\noindent\textbf{Qualitative comparison with the baseline.}
We employ Semi-DETR as the baseline and analyze the impact of Query Refinement on the one-to-many assignment strategy, as indicated in Figure~\ref{fig:queries-results}. 
Sparse Semi-DETR generates more accurate and refined positive proposals for detecting small or hidden objects. Furthermore, our method significantly reduces the input queries to the decoder compared to Semi-DETR in the one-to-many assignment strategy.
\begin{table}[ht]
    \vspace{-5pt}
    \centering
    \footnotesize
    \begin{tabular}{c|l}
        \multirow{2}{*}{Approach } & Training time  \\
        & (min)  \\
        \hline
        Semi-DETR & 38.56 \\
        \rowcolor{gray!20} Sparse Semi-DETR  & \textbf{34.38}~~\red{+4.18} \\
    \end{tabular}
    \vspace{-8pt}
    \caption{This is the training time for 1k iterations in one-to-many assignment strategy.}
    \label{tab:ablation}
    %\vspace{-10 pt}
\end{table}
%Furthermore, our method significantly reduces the number of queries fed to the decoder compared to Semi-DETR in the one-to-many assignment strategy.  
As evidenced in Table~\ref{tab:ablation}, this refinement of queries has resulted in a training time reduction of 4.18 minutes on 1k iterations, amounting to a relative decrease of 10.84\%. To further compare our Sparse Semi-DETR with the baseline Semi-DETR, we visualize the predicted bounding boxes on test2017, trained on the COCO 10\% label data. In Figure~\ref{fig:result6} and Figure~\ref{fig:result7}, we plot the predicted bounding boxes in red, while green and blue boxes highlight the differences in the prediction of Semi-DETR and Sparse Semi-DETR. There are four general properties that we could observe in our demonstration.

\begin{enumerate}
    \item Firstly, Sparse Semi-DETR significantly improves the detection of small objects compared to Semi-DETR, primarily due to its advanced query refinement mechanism. As shown in Figure~\ref{fig:result6} (a), Sparse Semi-DETR is particularly beneficial for identifying small subjects such as birds, where Semi-DETR often struggles because of its inadequate query feature representation. By capturing refined details, Sparse Semi-DETR ensures more precise and reliable detection of these smaller objects, enhancing overall performance in object detection tasks.
    \item Secondly, for obscured objects, Sparse Semi-DETR provides a distinct advantage over Semi-DETR through its refined query mechanism as indicated in Figure~\ref{fig:result6} (b). It allows Sparse Semi-DETR to understand better details of partially hidden objects, which is often challenging for Semi-DETR due to its less robust query features. As a result, Sparse Semi-DETR achieves more precise detection of obscured objects, leading to improved performance in complex visual environments.
    \item Thirdly, Sparse Semi-DETR exhibits a significant advantage in removing duplicate predictions after the second stage. It is because of a reliable pseudo-label filtering module that filters out some duplications and selects more accurate pseudo-labels. A notable example is the detection of cow objects, as shown in Figure~\ref{fig:result7} (a). While Semi-DETR tends to provide two predictions for the same object, Sparse Semi-DETR demonstrates remarkable proficiency in duplicate removal. 
    \item Fourthly, in the semi-supervised setting, Semi-DETR often faces challenges in accurately categorizing objects, even when the location is correctly identified. For example, Semi-DETR labels a 'remote' object as a 'cell phone' despite accurately providing its location as indicated in Figure~\ref{fig:result7} (b). This misclassification often arises from a disparity between the features used for object detection (regression) and those used for classification. In contrast, Sparse Semi-DETR stands out by adeptly distinguishing between closely related categories. It leverages its innovative attention and similarity module, which dynamically selects the most relevant features for each task, ensuring a more unified and accurate performance in both classification and localization.
\end{enumerate}


\end{document}